\documentclass[conference]{IEEEtran}
\IEEEoverridecommandlockouts
\usepackage{cite}
\usepackage{amsmath,amssymb,amsfonts}
\usepackage{algorithmic}
\usepackage{graphicx}
\usepackage{textcomp}
\usepackage{xcolor}
\def\BibTeX{{\rm B\kern-.05em{\sc i\kern-.025em b}\kern-.08em
    T\kern-.1667em\lower.7ex\hbox{E}\kern-.125emX}}
    
\usepackage{amsmath}
\usepackage{xspace}
\usepackage{booktabs}
\usepackage[shortlabels]{enumitem}
\usepackage{placeins}
\usepackage{algorithm}
\usepackage{algorithmic}
\usepackage{subfigure}
\usepackage{footnote}
\makesavenoteenv{tabular}
\usepackage[hidelinks]{hyperref}
    
\newcommand{\sysname}{\texttt{Snoopy}\xspace}

\newcommand{\RR}{\mathbb{R}}

\newcommand{\func}[3][f]{#1\colon #2\rightarrow #3}

\newcommand{\E}{\mathbb{E}}

\newcommand{\argmax}{\mathop{\mathrm{arg\,max}}}

\newcommand{\RH}{\widehat{R}}

\newtheorem{thm}{Theorem}[section]

\newtheorem{lem}[thm]{Lemma}

\def \cY {\mathcal Y}
\def \cX {\mathcal X}

\def \cD {\mathcal D}
\def \cL {\mathcal L}

\def \cF {\mathcal F}
\def \cO {\mathcal O}

\def \PP {\mathbb P}

\def \tcX {\widetilde \cX}

\def \tY {\widetilde Y}
\def \ty {\widetilde y}

\newenvironment{proofof}[1]{\indent{\scshape Proof of #1}:~~}{}
\newenvironment{proofcap}{\indent{\scshape Proof}:~~}{}

\begin{document}

\title{
Automatic Feasibility Study via Data Quality Analysis for ML: A Case-Study on Label Noise
\thanks{* The first three authors contributed equally to this paper.}
}

\author{\IEEEauthorblockN{Cedric Renggli$^{*,\dagger}$,~~Luka Rimanic$^{*,\dagger}$,~~Luka Kolar$^{*,\dagger}$,~~Wentao Wu$^\S$,~~Ce Zhang$^\dagger$}\\
\IEEEauthorblockA{$^{\dagger}$ETH Zurich,~~$^\S$Microsoft Research\\
  $^\dagger$\{cedric.renggli, luka.rimanic, luka.kolar, ce.zhang\}@inf.ethz.ch,~~$^\S$wentao.wu@microsoft.com}
}

\maketitle

\begin{abstract}
In our experience of working with domain experts who are using today's AutoML systems, a common problem we encountered is what we call ``\textit{unrealistic expectations}'' -- when users are facing a very challenging task with a noisy data acquisition process, while being expected to achieve startlingly high accuracy with machine learning (ML). Many of these are predestined to fail from the beginning.
In traditional software engineering, this problem is addressed via a \textit{feasibility study}, an indispensable step before developing any software system.
In this paper, we present \sysname, with the goal of supporting data scientists and machine learning engineers performing a \emph{systematic} and \textit{theoretically founded} feasibility study \emph{before} building ML applications. 
We approach this problem by estimating the \emph{irreducible error} of the
underlying task, also known as the Bayes error rate (BER), which stems from data quality issues in datasets used to train or evaluate ML models. We design a practical Bayes error estimator that is compared against baseline feasibility study candidates on 6 datasets (with additional real and synthetic noise of different levels) in computer vision and natural language processing. Furthermore, by including our systematic feasibility study with additional signals into the iterative label cleaning process, we demonstrate in end-to-end experiments how users are able to save substantial labeling time and monetary efforts.
\end{abstract}

\begin{IEEEkeywords}
Feasibility Study for ML, Data Quality for ML
\end{IEEEkeywords}

\section{Introduction}
\label{sectionIntro}

Modern software development is typically guided by software engineering principles that have been developed and refined for decades~\cite{van2008software}.
Even though such principles are yet to come to full fruition regarding the development of machine learning (ML) applications, in recent years we have witnessed a surge of work focusing on ML usability through supporting efficient ML systems~\cite{mulee-parameterSever,mllib, zaharia2018accelerating}, enhancing developer's productivity~\cite{bergstra2013hyperopt, baylor2017tfx,li2017hyperband}, and supporting the ML application development process itself~\cite{vartak2016modeldb, Northstar, polyzotis2019data, nakandala2019incremental, nakandala2020cerebro, hubis2019quantitative,CI, renggli2020model}. 

\paragraph*{\underline{Calls for a Feasibility Study of ML}} 
In this paper, we focus on one specific 
``failure mode'' 
that we frequently witness whilst
working with a range of domain experts,
which we call
``{\em unrealistic expectations}.''
Unlike classical software artifacts, the quality of ML models (e.g., its accuracy) is often a reflection of the \textit{data quality} used to train or test the model.
We regularly see developers that work on challenging tasks with
a dataset that is \textit{too} noisy to meet 
the unrealistically high
expectations on the accuracy 
that can be achieved with ML --- such a project is predestined to fail. Ideally, problems of this type should be
caught \emph{before} a user commits
significant amount of resources to
train or tune ML models.

In practice, if this were done by a \textit{human ML consultant},
she would first analyze the representative dataset for the defined task and assess the \textit{feasibility} of the target accuracy --- if the target is not achievable, one can then explore alternative options by refining the dataset, the acquisition process, or investigating different task definitions. Borrowing the term from classic software engineering, we believe that such a \textit{feasibility study} step is crucial to the usability of future ML systems for application developers.
In this paper, we ask: \textit{Can we provide some systematic and theoretically understood guidance for this feasibility study process?}

\paragraph*{\underline{Quantitative Understanding of ``Data Quality for ML''}} 
Data quality, along with its cleaning, integration, and acquisition, is
a core data management problem that has been intensively studied 
in the last few decades~\cite{cong2007improving, chiang2008discovering, sadiq201120, fan2015data, abedjan2016detecting, schelter2018automating, ilyas2019data}.
Agnostic to ML workloads, the data management community has been conducting a flurry of work aimed at understanding and quantifying data quality issues~\cite{wang1996beyond, strong1997data,scannapieco2002data,batini2009methodologies}.
In addition to these fundamental results, the presence of 
an ML training procedure as a downstream task over data
provides both challenges and opportunities.
On the one hand, systematically mapping these challenges to ML model quality issues is largely missing (with the prominent exceptions of ~\cite{krishnan2016activeclean, ghorbani2019data, wu2020complaint}
together with some of our own previous efforts~\cite{li2019cleanml, jia2019towards, jia2019efficient, karlavs2020nearest}).
On the other hand, the ML training procedure 
provides a quantitative metric to measure precisely 
the utility of data, or its quality. In this paper, we 
take one of the early steps in this direction and ask: 
``\textit{Can we quantitatively map the quality requirements of
a downstream ML task to the requirements of the data quality of
the upstream dataset?}'' 

\paragraph*{\underline{The Scope and Targeted Use Case}}
As one of the first attempts towards understanding this fundamental problem, this paper
by no means provides a complete solution. Instead, we have a very specific 
application scenario in mind for which we develop a 
deep understanding both theoretically
and empirically. 
Specifically, we focus on the case in which 
a user has access to a dataset $\mathcal{D}$, 
large enough to be representative
for the underlying task. 
The user is facing the following question:
{\em 
Is my current data artefact $\mathcal{D}$ clean/good enough
for \underline{some} ML models
to reach a target accuracy $\alpha_\text{target}$? 
} If the answer to this question is ``Yes'',
the user can start expensive AutoML runs and hopefully
can
 find a model that can reach $\alpha_\text{target}$;
otherwise, it would be better for the user 
to improve the \textit{quality} (via
data cleaning for example) of the data artefact $\mathcal{D}$
before starting an AutoML run which is ``doomed to disappoint''.
We call the process of answering 
this question a ``\textit{feasibility study}''.
Our main goal is to derive a strategy for the feasibility study that is \textbf{(i)} \textit{informative and theoretically justified}, \textbf{(ii)} \textit{inexpensive}, and  \textbf{(iii)} \textit{scalable}.  

Such a process can be useful in many scenarios. In this paper we develop a fundamental building block of the feasibility study and evaluate it focusing on
one specific use case as follows --- the dataset $\mathcal{D}$
is noisy in its labels, probably caused by (1) 
the inherent noise of the data collection 
process such as \textit{crowd sourcing}~\cite{compton2012developing, gadiraju2015understanding,checco2018all,sun2020improving},
or (2) \textit{bugs} in the data 
preparation pipeline (which we actually see quite often in practice). The user has a target
accuracy $\alpha_\text{target}$ and can spend time and 
money on two possible
operations: (1) manually clean up some 
labels in the dataset, or (2) 
find and engineer
suitable ML models, manually or automatically.

\paragraph*{\underline{Challenges of the Strawman}}
There are multiple strawman strategies, each 
of which has its own challenge.
A natural, rather trivial approach 
is to run a \textit{cheap proxy model}, e.g.,
logistic regression,
to get an accuracy $\alpha_\text{proxy}$,
and use it to produce an 
estimator $\alpha_\text{est}=c\cdot\alpha_\text{proxy}$, say, for some $c \in [1, 1/\alpha_\text{proxy}]$. The challenge of this approach is to pick a universally good constant $c$, which depends not only on
the data but also on the cheap proxy model. It is important, but often 
challenging, to provide a principled and theoretically justified way of adjusting the gap between $\alpha_\text{proxy}$ and  $\alpha_\text{est}$.

An alternative approach would be to simply fire up 
an AutoML run that systematically looks at various configurations
of ML models and potentially neural architectures.
Given enough time and resources, this could converge closely to the best possible accuracy that one can achieve
on a given dataset; nevertheless, this can be very expensive 
and time consuming, thus
might not be suitable for a quick feasibility study.

\paragraph*{\underline{Feasibility Study: Theory vs. Practice}} 
The main challenges of the strawman approaches motivate us to 
look at this problem in a more principled way.
From a theoretical perspective, 
our view on feasibility 
study is not new, rather it connects to 
a decades-old ML 
concept known as the 
Bayes error rate (BER)~\cite{Cover1967-zg},
\textit{the ``irreducible error'' of a given task}
corresponding to the error rate of the
\textit{Bayes optimal classifier}. In fact, all factors leading to an increase of the BER can be mapped to classical data quality dimensions (e.g., ``label noise'' to ``accuracy'', or ``missing features '' to ``completeness'')~\cite{renggli2021data}.
Estimation of the BER has been studied 
intensively for almost half a century by the ML community~\cite{Fukunaga1975-zx,Devijver1985-gs,Fukunaga1987-lw,buturovic1992improving,Pham-Gia2007-xa,Berisha2016-yl, sekeh2020learning}. Until recently, most, if not all, of these works are mainly theoretical, evaluated on either synthetic and/or very small datasets of often small dimensions. 
Over the years, we have been conducting a series of work 
aimed at understanding the behavior of these BER estimators on larger scale, real-wold datasets.
This paper builds on two of these efforts, notably (1) a framework to  compare BER estimators on large scale real-world datasets with unknown true BER~\cite{renggli2021evaluating}, and (2) new convergence bounds for a simple BER estimator on top of pre-trained transformations~\cite{rimanic2020convergence}.
Guided by the insights and theoretical understanding we gained from these prior works, which we treat as preliminaries and do not see them as a technical contribution of this work, we ask the following non-trivial questions:
\begin{quote}
\em Q1. How to use estimations of the BER for the purpose 
of systematic feasibility study for ML?
\end{quote}
\begin{quote}
\em Q2. How can we build a scalable system to make 
decades of theoretical research around the BER practical and feasible
on real-world datasets?
\end{quote}

\paragraph*{\underline{Summary of Contributions}} 
We present \sysname --- a fast, practical and systematic feasibility
study system for machine learning. 
We make three technical contributions.

\paragraph*{\bf C1. Systems Abstractions and Designs}
In \sysname, we model the problem of feasibility study as estimating a \emph{lower bound} of
the BER.
Users provide \sysname with a dataset representative for their ML task along with a target accuracy. The system then outputs a binary signal assessing whether the target accuracy is realistic or not.
Being aware of failures (false-positives and false-negatives) in the binary output of our system, which we carefully outline and explain in this paper, we support the users in deciding on whether to ``trust'' the output of our system by providing additional numerical and visual aids. 
The technical core is a
practical BER estimator.
We propose a simple, but novel
approach, which consults 
a collection of different 
BER estimators based on a 1NN estimator inspired by Cover and Hart\cite{Cover1967-zg}, built on top of a collection of pre-trained feature transformations, and aggregated through taking
the minimum. 
We provide a theoretical analysis
on the regimes under which this aggregation 
function is justified.

\paragraph*{\bf C2. System Optimizations}
We then describe the implementation of 
\sysname, with optimizations that
improve its performance.
One such optimization is the \emph{successive-halving} algorithm~\cite{jamieson2016non}, a part of the textbook Hyperband
algorithm~\cite{li2017hyperband}, to balance the resources
spent on different estimators. This 
already outperforms naive approaches significantly.
We further improve on this method by taking 
into consideration the convergence 
curve of estimators, fusing it into a new variant of successive-halving.
Moreover, noticing the iterative 
nature between
\sysname and the user, we 
take advantage of the property 
of kNN 
classifiers and implement 
an incremental version 
of the system. For scenarios 
in which a user cleans some labels, \sysname is able to 
provide real-time feedback (0.2 ms for 10K test samples and 50K training samples).

\paragraph*{\bf C3. Experimental Evaluation}
We perform a thorough experimental evaluation of \sysname on 6 well-known datasets in computer vision and text classification against the baselines that use cheap and expensive proxy models. We show that \sysname consistently outperforms the cheap, and matches the expensive strategy in terms of predictive performance for synthetic and natural label noise, whilst being computational much more efficient than both approaches. In an end-to-end use-case, where noisy datasets are iteratively cleaned up to a fraction required to achieve the target, our system enables large savings in terms of overall cost, especially in cheap label-cost regimes. In label-cost dominated regimes (i.e., large label costs or cheap compute costs), our system adds little to no overhead compared to the baselines. Furthermore, by exploring the regimes in which \sysname fails to provide a correct answer, we show the benefits of additional signals given to the user.

\paragraph*{\underline{Limitations}} In this paper
we focus on the challenging endeavor of estimating the \emph{irreducible error} for the task definition and data acquisition process, originating from data quality issues. We focus on label noise, representing one of the most prominent source for non-zero irreducible error. The exploration of other aspects of poor data quality, such as noisy or incomplete features, are left as future work.
We by no means provide a conclusive solution to prevent unrealistic or very costly endeavors of training ML models with finite data. Rather, we view our contribution
as a first step towards a practical treatment of this problem, which is the key for enabling a systematic feasibility study for ML. 
Concretely, we focus on classification tasks which, compared to other ML tasks, benefit of a solid theoretical understanding of the irreducible error and ways of estimating it.
As a result, in Section~\ref{secPreliminaries} we carefully describe limiting assumptions on the data distributions, as well as failure causes and failure examples, presented in Sections~\ref{sectionSystemDesign} and~\ref{sec:experiments} respectively, hoping that this can stimulate future research from the community.

\paragraph*{\underline{Future Extension}}

The feasibility study functionality targeted in this paper is ideal for new ML projects designed to replace existing ``classical'' code with certain accuracy. Nothing prevents data scientists and ML engineers to use \sysname prior to any batch trained ML model though. This is particularly appealing in the context of data-centric AI, where the signal can be used to understand the impact of data actions (e.g., cleaning labels). For stream-based or continual learning there are some extra challenges. First, the window of data should typically be small in order to have a good representation of the current distribution, which renders an accurate estimation of the BER challenging. Secondly, it is unclear how a BER estimator can be designed to cope with adversarial examples. Both aspects represent interesting lines of future research. Finally, when understanding the impact of distributional drift, the test accuracy of a fixed model is typically inspected. Designing drift-aware BER estimator could to detect such a drift for any model on a distributional level is left as future work.

\section{Preliminaries}
\label{secPreliminaries}

In this section, we give a short overview over the technical terms and the notation used throughout this paper.
Let $\cX$ be the feature space and $\cY$ be the label space, with $C\!=\!|\cY|$. Let $X\in\cX,Y\in\cY$ be random variables. Let $p(X,Y)$ be their joint distribution, often simplified by $p(x,y)\!=\!p(X\!=\!x,\!Y\!=\!y)$. We define $\eta_y(x)\!=\!p(y|x)$ when $C>2$, and $\eta(x)\!=\!p(1|x)$ when $C\!=\!2$, assuming $\cY=\{0, 1\}$. 

\paragraph*{\underline{Bayes Error Rate}} \emph{Bayes optimal classifier} is the classifier that achieves the lowest error rate among all possible classifiers from $\cX$ to $\cY$, with respect to $p$. Its error rate is called the \emph{Bayes error rate (BER)} and we denote it by $R_{X,Y}^*$, often abbreviated to $R_{X}^*$ when $Y$ is clear from the context. It can be expressed as $R_{X}^* = \E_{X} \big[ 1- \max_{y\in\cY} \eta_y(x) \big]$.

\paragraph*{\underline{k-Nearest-Neighbor (kNN) Classifier}}
Given a training set $\cD_n:=\{ (x_1,y_1),\ldots,(x_n,y_n)\}$ and a new instance $x$, let $(x_{\pi(1)}, \ldots, x_{\pi(n)})$ be a reordering of the training instances by their distances from $x$, based on some metric (e.g., Euclidean or cosine dissimilarity). 
The \emph{kNN classifier} $h_{n,k}$ and its \emph{$n$-sample error rate} $(R_X)_{n,k}$ are defined by 
\vspace{-1em}
\begin{align*} h_{n,k} (x) &= \argmax_{y\in\cY} \sum_{i=1}^{k} \mathbf{1}_{\{ y_{\pi(i)} = y\}},\\
    (R_X)_{n,k} &= \E_{X,Y} \mathbf{1}_{\{ h_{n,k}(X) \neq Y\}},
\end{align*}
respectively. The \emph{infinite-sample error rate} of kNN is given by $(R_X)_{\infty, k} = \lim_{n\rightarrow \infty} (R_X)_{n,k}$. Cover and Hart derived 
the following fundamental~\cite{Cover1967-zg},
and now well-known,
relationship between the nearest 
neighbor algorithm and the BER (under mild assumptions on the underlying probability distribution):
\begin{equation}\label{eqnCoverHart}
\begin{tiny}
(R_X)_{\infty,1 } \geq  R_{X}^* \geq \frac{(R_X)_{\infty, 1}}{1+\sqrt{1-\frac{C (R_X)_{\infty, 1}}{C-1}}}.
\end{tiny}
\end{equation}
Determining such a bound for $k>1$ and $C>2$ is still an open problem, and in this work we mainly focus on $k=1$. 

\begin{figure*}[t!]
\centering
\includegraphics[width=0.9\textwidth]{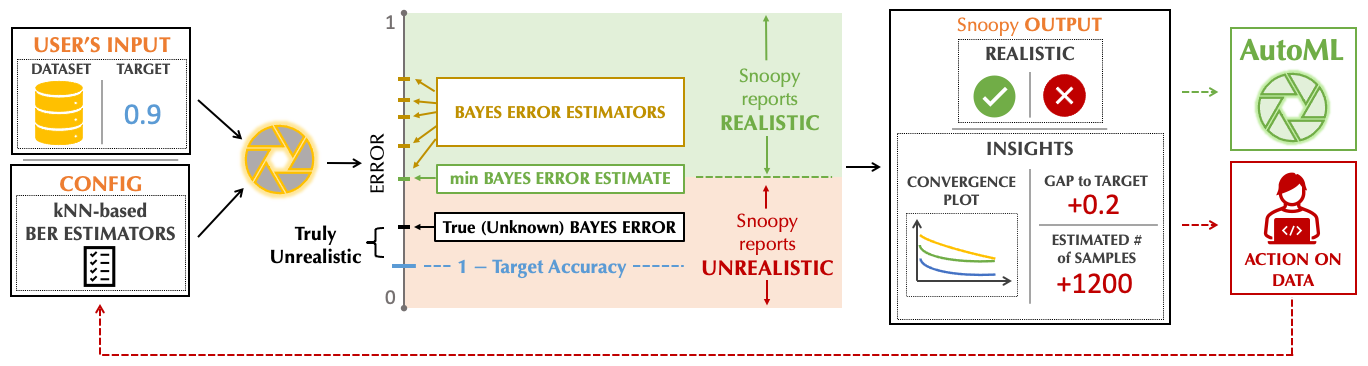}
\vspace{-1em}
\caption{Overview of \sysname: Given user's input in terms of a dataset and a target, the system consults various Bayes error estimators (yellow ticks), aggregates them by taking the minimum (green tick) and outputs its belief whether the target is realistic, together with the insights in terms of convergence plots, gap towards target accuracy, and estimated number of additional samples needed to reach the target accuracy.}
\label{fig:snoopy_overview}
\vspace{-1.5em}
\end{figure*}

\paragraph*{\underline{Bayes Error Estimation}} 
The task of estimating the BER, given a \emph{finite} representative dataset, is inherently difficult 
and has been investigated by the ML community for decades --- from Fukunaga's early work in 1975~\cite{Fukunaga1975-zx} to Sekeh et al.'s work in 2020~\cite{sekeh2020learning}. 
Existing BER estimators can be divided into three groups: \emph{density estimators} (\emph{KDE}~\cite{Fukunaga1987-lw}, \emph{DE-kNN}~\cite{Fukunaga1973-wg}), \emph{divergence estimator} (\emph{GHP}~\cite{sekeh2020learning}), \emph{kNN  classifier accuracy} (\emph{1NN-kNN}~\cite{Devijver1985-gs}, \emph{kNN-Extrapolation}~\cite{Snapp1996-fe}, 1NN inspired by~\cite{Cover1967-zg}). 

As mentioned earlier, we have been conducting a series of work 
in order to understand the theoretical and empirical behavior of deploying and comparing BER estimators on larger scale, real-world datasets, using powerful pre-trained embeddings.
This paper builds on these efforts~\cite{renggli2021evaluating, rimanic2020convergence} but treats them as 
preliminaries --- they provide
important insights into many decisions in our system, but they do not 
count as the technical contribution of this paper.
We next summarize these efforts and the gained insights.

\subsection{Evaluating Bayes Error Estimators on Real-World Datasets}

Evaluating the relative performance of BER
estimators on real-world dataset is far from trivial. In 
one of our previous endeavor~\cite{renggli2021evaluating}, we proposed \texttt{FeeBee}, a novel framework for evaluating BER estimators on real-world data. The key insights for building such a framework lies in the realization that evaluating BER estimators on a single point for tasks with unknown true BER is infeasible. Rather, one has to construct a series of points, for which the evolution of the BER is known. We do so by injecting uniformly distributed noise over the labels for different amounts of label noise and following the evolution of the BER through the following theoretical result.

\begin{lem}[From~\cite{renggli2021evaluating}]\label{lemLabelFlipping}
    Let $Y_\rho$ be a random variable defined on $\cY$ by setting $Y_\rho = Z \cdot U(\cY) + (1-Z) \cdot Y,$ where $U$ is a uniform variable taking values in $\mathcal{Y}$, and $Z$ is a Bernoulli variable with probability $0 \leq \rho \leq 1$, both independent of $X$ and $Y$. Then $R_{X, Y_\rho}^* = R_{X,Y}^* + \rho(1-1/C-R_{X,Y}^*).$
\end{lem}

The above lemma is sufficient in determining the strength of each BER estimator~\cite{renggli2021evaluating}. However, in order to apply a BER estimator in a system for a feasibility study in real-world datasets, where human annotators typically introduce more noise for classes which are harder to distinguish, we need to be able to go beyond uniform noise. Therefore, in Section~\ref{sec_noise_beyond_uniform} we provide a generalization that does not assume uniform label noise.

As a major finding of \texttt{FeeBee}, we established that the \emph{1NN}-based estimator is 
a powerful one --- on par or better than all other estimators when it comes to performance, whilst being highly scalable and insensitive to hyper-parameters~\cite{renggli2021evaluating}.
In this paper, \emph{1NN}-based estimator on top of a feature transformation is our default choice. For a fixed transformation $f$, and $n$-samples, it is defined by
\vspace{-1em}
\begin{equation}\label{eqn:1nn_estimator}
\begin{small}
\widehat{R}_{f(X),n} = \frac{(R_{f(X)})_{n, 1}}{1+\sqrt{1-\frac{C (R_{f(X)})_{n, 1}}{C-1}}}.
\end{small}
\end{equation}

\subsection{On Convergence of Nearest Neighbor Classifiers over Feature Transformations}

Since our estimator will combine the 1NN algorithm with pre-trained feature transformations, also called \textit{embeddings},
such as those publicly available, we need to understand the influence of such transformations. In our theoretical companion to this paper~\cite{rimanic2020convergence}, we provide a novel study of the behavior of a kNN classifier on top of a feature transformation, in particular its convergence rates on \emph{transformed} data, previously known only for the \emph{raw} data.\footnote{We restrict ourselves to $C=2$, as usual in theoretical results about the convergence rates of a kNN classifier.}
We prove the following theorem, recalling that a real-valued function $g$ is \emph{$L$-Lipschitz} if $|g(x) - g(x')| \leq L \| x - x'\|$, for all $x,x'$, and defining $\cL_{g}(f) := \mathbb{E}_X[(g \circ f)(X) - \eta(X)]^2.$

\begin{thm}[From~\cite{rimanic2020convergence}]\label{thm_theory_MainThm}
	Let $\cX \subseteq \RR^D$ and $\tcX\subseteq \RR^d$ be bounded sets, and let $(X,Y)$ be a random vector taking values in $\cX \times \{0,1\}$. Let $\func[g]{\tcX}{\RR}$ be an $L_g$-Lipschitz function. Then for all transformations $\func{\cX}{\tcX}$,\vspace{-0.2em}
	\begin{align}\label{eqnConvRatesThm}
	&\E_n \left[ (R_{f(X)})_{n,k}\right] - R_{X}^* \nonumber \\
	&\hspace{5mm}=\cO \left( \frac{1}{\sqrt{k}}\right) + \cO\Big( L_g \left( \frac{k}{n}\right)^{1/d} \Big) + \cO\left( \sqrt[4]{\cL_{g}(f)} \right).
	\end{align}
\end{thm}

Motivated by the usual architecture of trained embeddings, in Theorem~\ref{thm_theory_MainThm} one should think of $g$ as a softmax prediction layer with weights $w$, which allows taking $L_g = \| w\|^2$.   Equation~\ref{eqnConvRatesThm} shows that there is a trade-off between the improved convergence rates 
(in terms of $L_g$ and $d$) and the bias introduced by the transformation independent of kNN.

\section{Design of \sysname}
\label{sectionSystemDesign}

We next present the design of \sysname. A high-level overview of the workflow of our system is given in Figure~\ref{fig:snoopy_overview}.

\paragraph*{\underline{Functionality}}
\sysname~interacts with
users in a simple way.
The user provides an input
dataset that is representative for the classification 
task at hand, along with a target accuracy $\alpha_\text{target}$. The system then estimates 
the ``highest possible accuracy''
that an ML model can achieve, and outputs a binary signal --- \texttt{REALISTIC},
if the system deducts that
this target accuracy is achievable; \texttt{UNREALISTIC}, otherwise. We note that \sysname does not provide a model that can achieve that target, only its \textit{belief} on whether the target is achievable, using an \emph{inexpensive} process.
Furthermore, the goal of \sysname~is \textit{not} to provide a perfect answer on feasibility, but to give information that can guide and help with the decision-making process --- the signal provided 
by the system may as well be wrong, as we will discuss later.
The best possible accuracy is implicitly returned to the user in the form of the gap between target and projected accuracy (c.f., Section~\ref{sec:additional_guidance}).

\paragraph*{\underline{Interaction Model}}
The binary signal of \sysname given to a user is often correct, but not always. We now dive into the user's and \sysname's interaction upon receiving the signal.

\textit{The Case When \sysname Reports \texttt{REALISTIC}}: In general, one should trust the system's output when it reports the target to be realistic, and proceed with running AutoML. We note that wrongly reporting realistic can be a very costly mistake which any feasibility system should try to avoid. In theory, our system could also be wrong in that fashion, due to (1) a lower bound estimate based on the 1NN estimator by Cover and Hart~\cite{Cover1967-zg} that is known to be not always tight, or (2) the fact that the estimators are predicting asymptotic values. However, as presented in the next section, we construct our estimator in a theoretically justified way that aims at reducing such mistakes and in our experiments we do not observe this behavior. Even if this were the case, we expect (2) to be the dominating reason, in which case gathering more data for the task at hand and running AutoML on this larger dataset might very well confirm the system's prediction. 

\textit{The Case When \sysname Reports \texttt{UNREALISTIC}}: In this case, our experiments showed that the system's output is also trustworthy, but with more caution. Under reasonable computational resources\footnote{For instance, reproducing the state-of-the-art model performance for well-established benchmark datasets is often a highly non-trivial task.},
\sysname is often correct in preventing unrealistic expectations for a varying amount of both synthetic and natural label noise. Nonetheless, there are two possible reasons for making wrong predictions in this manner: (1) either the data is not representative enough for the task (i.e., users might need to acquire more data), or (2) the transformations applied in order to reduce the feature dimension, or to bring raw features into a numerical format in the first place (e.g., from text), increased the BER.\footnote{We have shown in our theoretical companion~\cite{rimanic2020convergence} that any deterministic transformation can only increase the BER.} We note that (1) and (2) are complementary to each other. Even though estimating the BER on raw features (if applicable) prevents (2) from happening, having ``better'' transformations can lower the number of samples required to accurately estimate the BER.
In an ideal world, one could rule out (1) by checking whether the BER estimator converged on the given number of samples. That is why \sysname provides insights in terms of convergence plots and finite-sample extrapolation numbers to help users understand the relation between increasing number of samples and BER estimate, giving insights into the \emph{source} of predicting \texttt{UNREALISTIC}, and increase the confidence in the prediction.

\subsection{Data Quality Issues and the BER}\label{sec_noise_beyond_uniform}

The power of the BER, and the reason that 
\sysname focuses on estimating this quantity, is that it provides a link connecting data quality to the performance
of (best possible) ML models. This link can be 
made more explicit, even in closed form, if we assume some noise model. We take one of the most 
prominent source of data quality issues, 
label noise, as 
an example, and illustrate this connection
via a novel theoretical analysis.

\paragraph*{\underline{Noise Model}}
We focus on a standard noise model: 
class-dependent label noise~\cite{wei2022learning}. We assume that we are given a noisy random variable $Y_\rho$ through a transition matrix $t$ with
\vspace{-0.5em}
\begin{equation}\label{eqn_transition_matrix}
t_{\ty,y} := \PP (Y_\rho = \ty \ | Y = y, X = x) = \PP (Y_\rho = \ty \ | Y = y),
\vspace{-0.5em}
\end{equation}
where the equality follows from the assumption that we are in the class-dependent label noise scenario, rather than in instance-dependent one. One can think of $\rho (y) = 1 - t_{y, y}$ to be the fraction of class $y$ that gets flipped. Let $y_x := \argmax_{y\in \cY} p(Y=y | x)$. We further assume that $y_x = \argmax_{y \in \cY} p_\rho (Y_\rho =y | x)$, meaning that the maximal label per sample $x$ is preserved after flipping, albeit possibly with lower probability (which then increases the BER). Our main result is the following theorem.

\begin{thm}\label{thm_flipping}
Let $Y_\rho$ be a random variable taking values in $\cY$ that satisfies (\ref{eqn_transition_matrix}). Then
\vspace{-0.5em}
\begin{align*}
    R_{X, Y_\rho}^* &= R_{X,Y}^* + \E_X [\rho (y_x) p(y_x | x)] - \sum_{y \neq y_x} \E_{X} \left[ t_{y_x, y} p (y | x) \right] 
\end{align*}
\end{thm}
One can prove Theorem~\ref{thm_flipping} using the law of total expectation, together with careful manoeuvring of the terms that involve the elements of the transition matrix.\footnote{We provide the full proof and additional discussion in the Appendix~\ref{app:label_flipping}.}
Setting $\rho(y) = \rho \cdot \left( 1- 1/C \right)$, for all $y\in \cY$, and $t_{y,y'} = \rho / C$, for all $y' \neq y$, recovers Lemma~\ref{lemLabelFlipping}, and one can further deduct the following valid bounds on the evolution of the BER:
\vspace{-0.5em}
\begin{align*}
    (1-s_{X,Y}) \min_{y}\rho(y) - s_{X,Y} \max_{y,y': y\neq y'} t_{y,y'} \leq \\
    R_{X,Y_\rho}^* \leq s_{X,Y} + \max_{y}\rho(y),
\end{align*}
where $s_{X,Y}$ denotes the error of state-of-the-art model. 

\paragraph*{\underline{Other Data Quality Dimensions}}
Whilst we assume that the BER for zero label noise is typically small, it does not have to be equal to zero (c.f.,~\cite{renggli2021data} for examples). Nonetheless, by estimating the BER, we implicitly quantify the data quality issues
along \textit{all} dimensions (e.g., missing features, or combinations of feature and label noise). Deriving alternative noise models to theoretically and empirically disentangling these factors is a challenging task and left as future work.

\section{Implementation}
\label{sectionSystemImplementation}

The core 
component of \sysname is a 
BER estimator, which 
estimates the irreducible error of a given task.
The key design decision of 
\sysname is to \emph{consult a
collection of BER estimators and
aggregate them in a meaningful way.}
More precisely, for a collection of feature transformations $\cF$, e.g., 
publicly available pre-trained feature transformations (or last-layer representations of pre-trained neural networks) on platforms like TensorFlow Hub, PyTorch Hub, and HuggingFace Transformers, we define our main estimator of the BER (on $n$ samples) using Equation~\ref{eqn:1nn_estimator} by
\vspace{-0.5em}
\[
\RH = \min_{f \in \cF}\widehat{R}_{f(X),n}.
\vspace{-0.5em}
\]
Finally, the system's output is
\vspace{-0.5em}
\begin{align*}
\texttt{REALISTIC}, &\hspace{3mm}\text{ if } \RH \leq 1- \alpha_\text{target}, \\
\texttt{UNREALISTIC}, &\hspace{3mm}\text{ otherwise.}
\vspace{-0.5em}
\end{align*}

\subsection{``Just a Lightweight AutoML System?''}
At first glance, our system
might seem like a 
``lightweight AutoML system,''
which runs a collection of 
fast models (e.g., kNN classifiers) and takes the minimum to 
get the best possible classifier accuracy. 
We emphasize the difference --- the accuracy of an AutoML system always corresponds to 
a concrete ML model that can achieve this accuracy; however,
a BER estimator 
does \textit{not} provide this concrete model.
That is, \sysname~does \textit{not} construct a model 
that can achieve $\RH$. This key difference between
AutoML and feasibility study makes the latter inherently more computationally efficient, with almost instantaneous re-running, which we will further illustrate with experiments in Section~\ref{sec:endtoend}.

\subsection{Theoretical Analysis}
\label{sectionTheoreticalAnalysis}

Given a collection of 1NN-based BER estimators over feature transformations, 
\sysname~aggregates them by taking the minimum. 
This seemingly simple aggregation rule is far from trivial, raising obvious questions --- \textit{Why can we aggregate BER estimators by taking the minimum?}
\textit{When will this estimator work well and when will it not?}

In order to mathematically quantify different regimes, we need a few simple definitions.
, which we illustrate in Appendix~\ref{app:justifications_min} of the supplementary material.
We define the \emph{asymptotic tightness} of our estimator for a fixed transformation $f$ as
\vspace{-0.5em}
\begin{equation}
\Delta_f = R_{f(X)}^* - \lim_{n\rightarrow \infty} \widehat{R}_{f(X),n}.
\vspace{-0.5em}
\end{equation}
Equation~\ref{eqnCoverHart} implies $\Delta_f \geq 0$. We define the corresponding \emph{transformation bias} by
\vspace{-0.5em}
\begin{equation}
\delta_f = R_{f(X)}^* -R_{X}^*,
\vspace{-0.5em}
\end{equation}
with $\delta_f \geq 0$ (by~\cite{rimanic2020convergence}).
Finally, the \emph{n-sample gap (of the estimator)} is given by
\vspace{-0.5em}
\begin{equation}
\gamma_{f,n} = \widehat{R}_{f(X),n} - \lim_{n\rightarrow \infty}\widehat{R}_{f(X),n},
\vspace{-0.5em}
\end{equation}
with $\gamma_{f,n} \geq 0$ in expectation (also by~\cite{rimanic2020convergence}).

The fundamental challenge lies in the fact that none of the three quantities above can be derived in practice: $\Delta_f$ is dependent on the underlying unknown distribution, $\delta_f$ is intractable for complex neural networks~\cite{rimanic2020convergence}, and $\gamma_{f,n}$ relies on the convergence of the estimator, which requires the number of samples to be exponential in the input dimension~\cite{Snapp1996-fe}, making it impossible to generalize to representations on real-world datasets. Nevertheless, the connection between the quantities, together with the empirical analysis from \cite{renggli2021evaluating} and Section~\ref{sec:experiments} of this work, allows us to define meaningful regimes next.

\textit{\underline{When is $\RH$ optimal?}} In other words, when does the transformation that yields the minimum outperform all the others? A sufficient condition
\footnote{We prove the condition to be sufficient in Appendix~\ref{app:justifications_min} in the supplementary material.} 
is given by
\vspace{-0.7em}
\begin{equation}
\label{eqnCase1}
\forall f \in \cF:\quad \delta_f + \gamma_{f,n} - \Delta_f \geq 0.
\vspace{-0.5em}
\end{equation}
If the sum of finite-sample gap and transformation bias (i.e., the normalized constants of the second and third terms in Equation~\ref{eqnConvRatesThm}) is larger than the asymptotic tightness of the estimator (i.e., the normalized constants of the first term in Equation~\ref{eqnConvRatesThm}) for all transformations, then all estimators yield a number larger than the true BER, and therefore the minimum can be taken. Intuitively, this means that all the curves in the convergence plot are above the true BER. We note that this has to include the identity transformation, where there is no transformation bias. If Condition~\ref{eqnCase1} holds, $\RH$ \textit{will not underestimate the BER}. The above trivially holds if for all $f\in \cF$ one has $\Delta_f = 0$. Note that any classifier accuracy (non-scaled, to be used as proxy) also trivially falls into this regime, although it is usually worse than $\RH$. Furthermore, the system is guaranteed to not predict YES when the target is unreachable, thus avoiding costly mistakes. If the system wrongly predicts UNREALISTIC, it is guaranteed that its predicted error is off by at most $\delta_f + \gamma_{f,n} - \Delta_f$.

We note that all empirical evidence in Section~\ref{sec:experiments} and in our companion work on BER evaluation framework~\cite{renggli2021evaluating} suggests that we are in this regime for reasonable label noise (e.g., less than 80\%) on a wide range of datasets and transformations.

\begin{figure}[t!]
\centering
\includegraphics[width=0.92\linewidth]{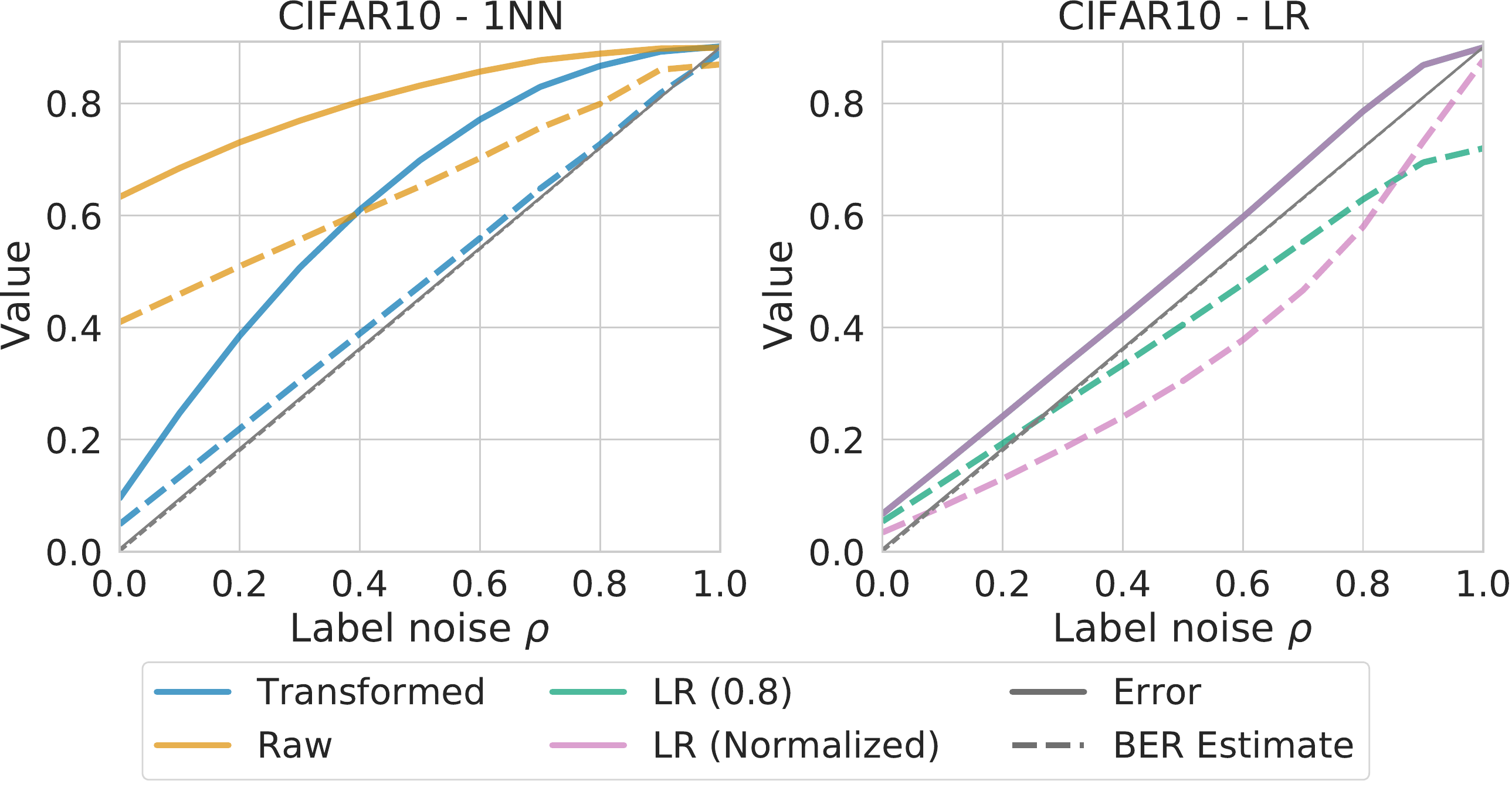}
\vspace{-1em}
\caption{Theoretical justifications: (Left) 1NN error and its estimator values for raw features and the best transformation. (Right) Scaling down the logistic regression error on the best transformation and normalizing it by plugging it into the 1NN estimator of Equations~\ref{eqn:1nn_estimator}. Notice that the solid green and solid pink line are identical in the right plot, leading to a dark purple line.}
\label{fig:feebee_for_snoopy_cifar10_1nn_and_lr}
\vspace{-1.5em}
\end{figure}

\textit{\underline{What if $\RH$ is not optimal?}} We distinguish two different cases in this regime. In the first one, we suppose that the suggested estimator $\RH_{X,\infty}$ of Cover and Hart~\cite{Cover1967-zg} on the \emph{raw} features performs well in the asymptotic regime, i.e. that $\Delta_{id}$ is small. In that case, a sufficient condition for $\RH$ to be at least as good as $\RH_{X,\infty}$ is $\Delta_f \leq \Delta_{id}$, for all $f \in \cF$.
Intuitively, this states that if all transformations do not increase the asymptotic tightness of the estimator by transforming the underlying probability distribution with respect to the raw distribution, taking the minimum over all transformations is no worse than running the estimator with 1NN on infinite samples. This condition can be seen empirically by inspecting the linear shape of the 1NN-based BER estimator values with increasing label noise for different transformations (c.f., Figure~\ref{fig:feebee_for_snoopy_cifar10_1nn_and_lr} on the left). One could weaken the condition for the finite-sample regime, resulting in a sufficient condition for $\RH$ to perform better than $\RH_{X,\infty}$:
\vspace{-0.5em}
\begin{equation}
\label{eqnCase2}
\forall f \in \cF:\quad \delta_f + \gamma_{f,n} - \Delta_f + \Delta_{id} \geq 0.
\vspace{-0.5em}
\end{equation}

For the second case, when $\RH$ performs poorly, we ask: \emph{What is the worst-case error of underestimation?} Using the fact that 1NN error is trivially above the true BER (c.f., left inequality of Equation~\ref{eqnCoverHart}), we can bound the difference of the 1NN-Based estimator value. In fact, the estimator value is at most the scaling factor of Equation~\ref{eqn:1nn_estimator} away from the true BER (i.e., $1/2$ for a binary classification problem).
However, our analysis and empirical verification reveals that our estimator of choice rarely ends up in the worst-case scenario. In fact, $\RH$ is usually the optimal choice and, when it is not, we often end up in the regime in which $\RH_{X,\infty}$ already performs well and $\RH$ outperforms it.

\paragraph*{\underline{Downscaling classifiers other than 1NN}}
As the worst-case error holds for scaling down any classifier accuracy,
one could be tempted to use a downscaled version (e.g., dividing by a constant $c>1$, or by plugging the error value in place of the 1NN error into the estimator of Equation~\ref{eqn:1nn_estimator}) of other classifiers as a proxy.
Contrary to the 1NN-based estimators, it is easy to show that for many datasets, any scaled version of a proxy model accuracy quickly falls into this worst-case regime (c.f.,  Figure~\ref{fig:feebee_for_snoopy_cifar10_1nn_and_lr} on the right, or Figure 4b in ~\cite{renggli2021evaluating}), supporting the challenges of the Strawman outlined in the introduction.

\subsection{Additional Guidance}\label{sec:additional_guidance}
To support users of \sysname in deciding whether to ``trust'' the output of the system, regardless of the outcome, additional information is provided. It comes in the form of (a) the estimated BER and, thus, the \emph{gap} between the projected accuracy and the target accuracy, (b) the convergence plots indicating the estimated BER value with respect to increased number of training samples over all deployed BER estimators (as illustrated in Figure~\ref{fig:snoopy_overview}), and (c) an additional estimate of the required number of additional samples to reach the target accuracy for the minimal transformation. Such an estimate is fairly non-trivial. Although Snapp et. al.~\cite{Snapp1996-fe} suggest how to approximate the kNN error by fitting a parametrized function to sampled data, the number of samples required to attain high confidence and accuracy is exponential in the feature dimension. This method is thus not practical for either finite-sample extrapolation, or estimating the BER, as shown in our companion work~\cite{renggli2021evaluating}. Instead, to support users of \sysname beyond purely visual aids, we approximate the estimate based on the 1NN error using a simple log-linear function~\cite{hashimoto2021model}
\vspace{-0.5em}
\begin{align}\label{eqnApproxExtrap}
\log\left((R_X)_{n,k}\right) \approx -\alpha \log\left(n\right) + C,
\vspace{-0.5em}
\end{align}
for two positive constants $\alpha$ and $C$. The idea of approximating the error is motivated by recent observations of scaling laws across different deep learning modalities 
~\cite{kaplan2020scaling, rosenfeld2020a}. Notice that Equation~\ref{eqnApproxExtrap} should only be used to extrapolate the convergence for a small number of additional data points. The function (i.e., the exponential of the righthand side of Equation~\ref{eqnApproxExtrap}) is known to converge to $0$, implying that regardless of the label noise or true BER, it will always underestimate the BER for a too large number of samples. We show the benefits and failures of using this approximation in Section~\ref{secExpAdditionalGuidance}.

\section{System Optimizations}
\label{sectionEfficientImplementation}

The suggested and theoretically motivated estimator from the previous section relies on the 
1NN classifier being evaluated on a possibly large collection of publicly available
\emph{pre-trained} feature transformations. We present optimizations that improve the performance, making it more scalable.

\paragraph*{\underline{Algorithm}} There are five computational steps involved:
\begin{enumerate}[label=\textbf{(\roman*)},align=parleft,left=0pt..2em]
    \item Take user's dataset with $n$ samples: features $X_1, X_2, \dots, X_n$ and labels $Y_1, Y_2, \dots, Y_n$. 
    \item For pre-defined $m$ transformations $\cF = \{f_1, f_2, \dots, f_m\}$, calculate the corresponding features for every sample in the dataset by applying all the transformations in $\cF$.
    \item For each feature transformation $j \in [m]$, calculate the 1NN classifier error $R_j=(R_{f_j(X)})_{n,1}$ on the transformed features $f_j(X_i)$, for all samples $i \in [n]$. 
    \item Based on the 1NN classifier error, derive the lower-bound estimates $\widehat{R}_{f_j(X),n}$ using Equation~\ref{eqn:1nn_estimator}. 
    \item Report the overall estimate $\widehat{R} = \min_{j \in [m]} \widehat{R}_{f_j(X),n}.$
\end{enumerate}

Note that the dataset is split into training samples and test samples. The test set is only used to estimate the accuracy of the classifier and is typically orders of magnitude smaller than the training set.
The quality of \sysname~depends heavily on the list of feature transformations that are fed into it. Since we take the minimum over all transformations in $\cF$, increasing the size of the set only \emph{improves} the estimator. On the downside, an efficient implementation is by no means trivial with an ever-increasing number of (publicly) available transformations.

\paragraph*{\underline{Computational Bottleneck}} When analyzing the previously defined algorithm, we realize that the major computational bottleneck comes from transforming the features. Especially when having large pre-trained networks as feature extractors, running inference on large datasets,
in order to get the embeddings, can be very time-consuming and result in running times orders of magnitude larger than the sole computation of the 1NN classifier accuracy. More concretely, given a dataset with $n$ samples and $m$ feature transformations, the worst case complexity is $\cO(m n)$, which highlights the importance of providing an efficient version of the algorithm.

\paragraph*{\underline{Multi-armed Bandit Approach}} Inspired by ideas for efficient implementation of the nearest-neighbor search on hardware accelerators~\cite{johnson2019billion}, running inference on all the training data for all feature transformations simultaneously is not necessary. Rather, we define a streamed version of our algorithm by splitting the steps (ii) to (iv) into iterations of fixed batch size per transformation.
This new formulation can directly be mapped to a \textit{non-stochastic best arm identification} problem, where each arm represents a transformation.
The successive-halving algorithm~\cite{jamieson2016non}, which is invoked as a subroutine inside the popular Hyperband algorithm~\cite{li2017hyperband}, is designed to solve this problem efficiently. We can summarize the idea of successive-halving as follows: Uniformly allocate a fixed initial budget across all transformations and evaluate their performance. Keep only the better half of the transformations, and repeat this until a single transformation remains.
The full algorithm is explained in the supplementary material in Appendix~\ref{app_SH_tangents}.

\begin{figure}[t!]
\centering
\includegraphics[width=0.62\linewidth]{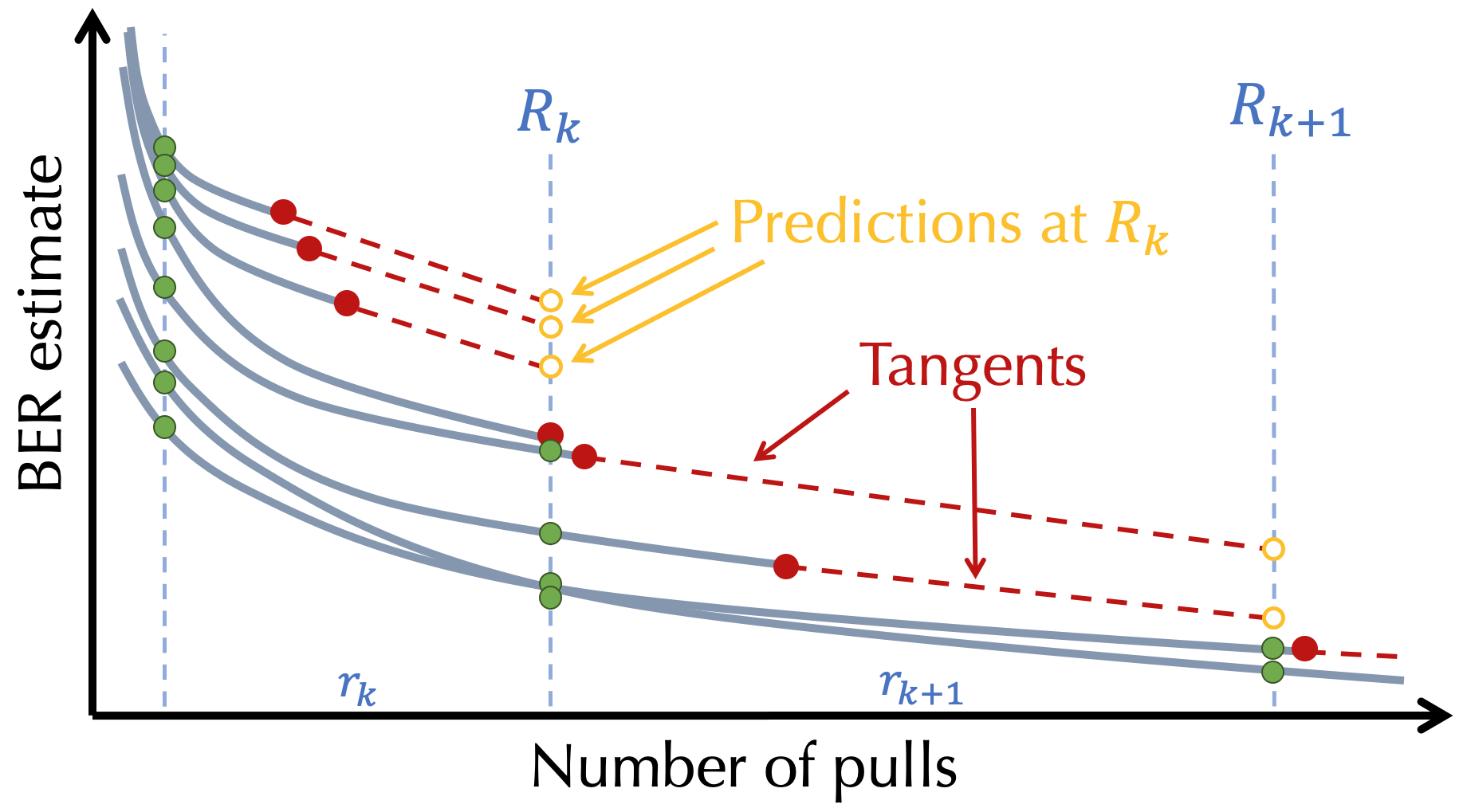}
\vspace{-1em}
\caption{Improved Successive-Halving: At each point for each convergence curve we construct a tangent and check whether there are more than half of the remaining curves that are better than the tangent.}
\label{fig:improved_SH}
\vspace{-1.5em}
\end{figure}

\paragraph*{\underline{Improved Successive-Halving}} We develop a variant of successive-halving that further improves the performance. The main idea comes from observing the \emph{convergence curve} of a kNN classifier. We know that under some mild assumptions, the kNN error decreases as a function of $n^{-2/d}$, where $n$ is the number of samples~\cite{snapp1991asymptotic}. Therefore, we can assume that the convergence curve is decreasing and convex. This allows us to predict a simple lower bound for the convergence curve at the end of each step -- using the tangent through the curve at the last known point, as illustrated in Figure~\ref{fig:improved_SH}. If the tangent line at the end point is worse than half of the remaining curves at the current point, the curve will not proceed to the next round. To simplify the implementation, we approximate the tangent by a line through the two last-known values of the convergence curve and develop a variant of successive-halving that uses this as a stopping condition. An important property of our improvement is that the remaining transformations after each step are the same as the ones from the original successive-halving, which implies that all theoretical guarantees of successive-halving still hold.

\paragraph*{\underline{Parameters of Successive-Halving}} We eliminate the dependency on the initial budget by implementing the \textit{doubling-trick} (cf. Section 3 in~\cite{jamieson2016non}). The batch size of the iterations has a direct impact on the performance and speedup of the algorithm. This is linked to properties of the underlying hardware and the fact that approximating the tangent for points that are further apart becomes less accurate. Hence, we treat the batch size as a single hyper-parameter, which we tune for all transformations and datasets.

\paragraph*{\underline{Efficient Incremental Execution}}
For the specific scenario of incrementally cleaning labels until a target accuracy is reachable, we provide a simple yet effective optimization that enables re-running \sysname almost instantly. After its initial execution, the system keeps track of the label of a single sample per test point -- its nearest neighbor. As cleaning labels of test or training samples does not change the nearest neighbor, calculating the 1NN accuracy after cleaning any training or test samples can be performed by iterating over the test set exactly once, thus, providing real-time feedback.

\begin{table}[t]
\caption{Datasets and SOTA performances.\label{tbl:datasets}}
\vspace{-1em}
\centering
\begin{tabular}{ lrrrr } 
\toprule
Name  & Classes $C$ & Train / Test Samples & SOTA \%\\
\midrule
MNIST  & 10 & 60K / 10K & 0.16~\cite{byerly2020branching} \\ 
CIFAR10  & 10 & 50K / 10K & 0.63~\cite{kolesnikov2019large}\\ 
CIFAR100  & 100 & 50K / 10K & 6.49~\cite{kolesnikov2019large}\\ 
\midrule
IMDB  & 2 & 25K / 25K & 3.79~\cite{yang2019xlnet}\\ 
SST2  & 2 & 67K / 872 & 3.2~\cite{yang2019xlnet}\\ 
YELP  & 5 & 500K / 50K & 27.80~\cite{yang2019xlnet}\\
\bottomrule
\end{tabular}
\vspace{-2em}
\end{table}

\section{Experiments}
\label{sec:experiments}

We now present the results of our empirical evaluation by describing the benefits of performing a feasibility study in general, and using the binary output of \sysname over other baselines. We focus on a specific use-case scenario motivated in the introduction. We also show how the additional guidance can increase trust in the binary signal. We then analyze the generalization properties of our system on certain vision tasks and conclude this section by performing a detailed performance analysis of \sysname. The code of \sysname is available via \url{https://github.com/easeml/snoopy}, whereas the code to reproduce the results can be found  under~\url{https://github.com/DS3Lab/snoopy-paper}.

\begin{figure*}[t!]
\centering
\includegraphics[width=0.92\textwidth]{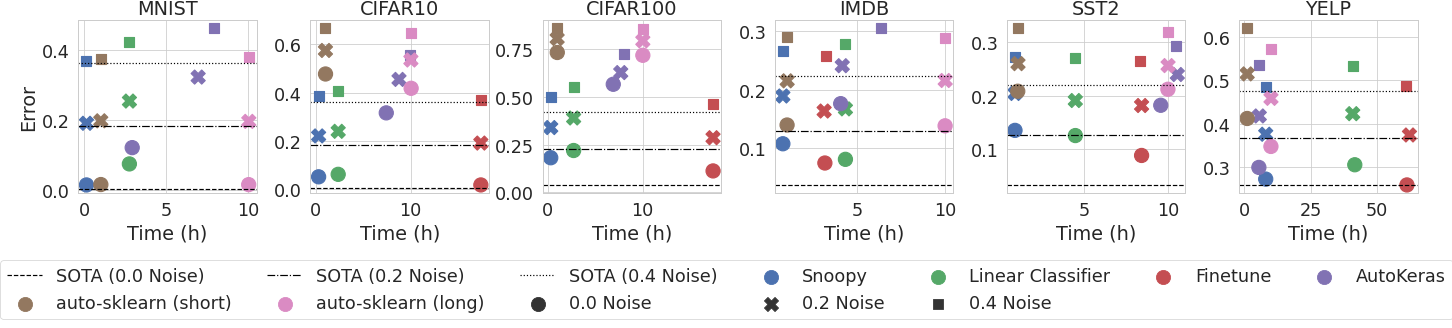}
\vspace{-1em}
\caption{Error Estimations vs. Time on three synthetics noise levels. The dashed horizontal lines represent the expected increase of the SOTA using Lemma~\ref{lemLabelFlipping}.}
\label{fig:snoopy_vs_baselines}
\vspace{-1.5em}
\end{figure*}

\subsection{Experimental Setup}

\paragraph*{\underline{Datasets}}

We perform the evaluation on two data modalities that are ubiquitous in modern machine learning and are accompanied by strong state-of-the-art (SOTA) performances summarized in Table~\ref{tbl:datasets}. Implicitly, a strong SOTA yields a low natural BER (i.e., originating from all data quality dimensions). The first group consists of visual classification tasks, including CIFAR10~\cite{dosovitskiy2020image}, CIFAR100~\cite{foret2020sharpness}, and MNIST~\cite{byerly2021no}. The second group consists of standard text classification tasks, where we focus on IMDB, SST2, and YELP~\cite{yang2019xlnet}. We remark that the SOTA values for SST2 and YELP are provided on slightly different sizes of training sets.

\begin{table}[t]
\caption{CIFAR-N datasets statistics.\label{tbl:dataset_cifar-n} The variable $t_{i,j}$ refers to an element of the noise transition matrix $t$ (c.f., Section~\ref{sec_noise_beyond_uniform}).}
\vspace{-1em}
\centering
\begin{tabular}{@{}lcccc@{}}
\toprule
Dataset         & Noise & $\max_{i} t_{i,i}$ & $\min_{i} t_{i,i}$ & $\max_{i\neq j} t_{i,j}$  \\ \midrule
CIFAR10-Aggre   & 9\%      & 17\%           & 3\%   & 10\%   \\
CIFAR10-Random1 & 17\%     & 26\%           & 10\%  & 23\%   \\
CIFAR10-Random2 & 18\%     & 26\%           & 10\%  & 23\%   \\
CIFAR10-Random3 & 18\%     & 26\%           & 10\%  & 23\%   \\
CIFAR100-Noisy  & 40\%     & 85\%           & 8\%   & 31\% \\ \bottomrule
\end{tabular}
\vspace{-2em}
\end{table}

We mainly focus our study on datasets with noisy labels. 
The ML community usually works on high-quality, noise-free benchmark datasets. As an exception, Wei et. al.~\cite{wei2022learning} published different noisy variants of the popular CIFAR datasets, called CIFAR-N. The noise levels vary between 10\% and 40\% (c.f., Table~\ref{tbl:dataset_cifar-n}). The datasets are provided with their noise transition matrix, allowing us to use the bounds derived from Theorem~\ref{thm_flipping}. The assumption therein corresponds to the each diagonal element being the maximal value per row, which is given for all datasets.
Supported by our theoretical understanding of the impact of label noise on the BER and its evolution, we also synthetically inject uniform label noise for 20\% and 40\% of the label into all six datasets from Table~\ref{tbl:datasets}.

\paragraph*{\underline{Feature Transformations}}

We compile a wide range of more than 15 different feature transformations per data modality, such as PCA and NCA~\cite{wu2018improving}, as well as state-of-the-art pre-trained embeddings. The pre-trained feature transformations are taken from public sources such as TensorFlow Hub, 
PyTorch Hub, and HuggingFace, whereas PCA and NCA are taken from scikit-learn\footnote{TensorFlow Hub: \url{https://tfhub.dev}, PyTorch Hub: \url{https://pytorch.org/hub}, HuggingFace: \url{https://huggingface.co/models/} and scikit-learn: \url{https://scikit-learn.org/}}. The pre-trained embeddings can either be directly accessed via the corresponding source, or have to be extracted from the last-layer representations of pre-trained neural networks.
More details about the transformations supported, for each modality individually, can be found in Appendix~\ref{secAppendixTransormations}.

\paragraph*{\underline{Settings of \sysname}} When running \sysname, we define the time needed to reach the lowest 1NN error across all embeddings based on multiple independent runs as described in Section~\ref{secAccuracyAndEfficiency}. These runtimes include the 1NN computation and running inference on a single GPU, with the latter being the most costly part, particularly for large NLP models. In the end-to-end experiments, when re-running \sysname after having restored a fixed portion of the synthetically polluted labels (set to 1\% of the dataset size), we use the fact that the ``best'' embedding did not change and, therefore, no additional inference needs to be executed.

We compare with a diverse set of baselines 
that estimate the BER: (i) training a logistic regression (LR) model on top of all pre-trained transformations, (ii) running AutoKeras, and (iii) fine-tuning a state-of-the-art (SOTA) foundation model~\cite{bommasani2021opportunities} for each data modality.

\begin{figure*}[t!]
\centering
\includegraphics[width=0.92\textwidth]{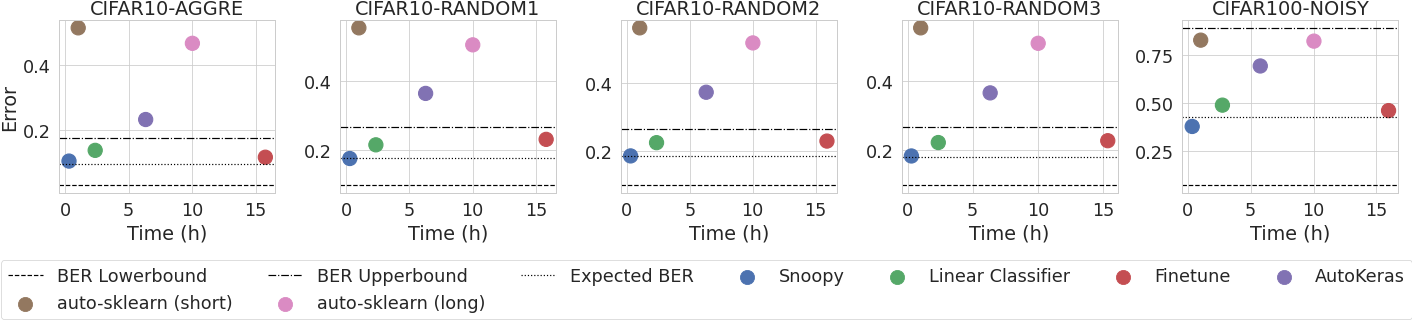}
\vspace{-1em}
\caption{Error Estimations vs. Time on multiple real noisy variant of CIFAR10 and CIFAR100 (c.f., Table~\ref{tbl:dataset_cifar-n}). The dashed horizontal lines represent bounds and the expected increase of the SOTA using Theorem~\ref{thm_flipping}.}
\label{fig:snoopy_vs_real_baselines}
\vspace{-1.5em}
\end{figure*}

\paragraph*{\underline{Baseline 1: LR Models}} As mentioned before, when training the logistic regression models we assume that the representations for all the training and test samples are calculated in advance exactly once. In the end-to-end experiments, after having restored the same fixed portion of labels (i.e., 1\% of test and train samples), re-training the LR models does not require any inference. We train all LR models on a single GPU using SGD with a momentum of 0.9, a mini-batch size of 64 and 20 epochs. We select the minimal test accuracy achieved over all combinations of learning rate in $\{0.001, 0.01, 0.1\}$ and $L_2$ regularization values in $\{0.0, 0.001, 0.01\}$. We calculate the average time needed to train a LR based on the best transformation, without label noise, on all possible hyper-parameters. The hyper-parameter search was conducted 5 independent times for any value of randomly injected noise.

\paragraph*{\underline{Baseline 2: AutoML Systems}} To mimic the use of an AutoML systems on a single GPU without any prior dataset-dependent knowledge, we run AutoKeras with the standard parameters of a maximum of 100 epochs and 2 trials on top of all datasets. We additionally run auto-sklearn with two different configurations to simulate a short execution time (max 1 hour), and a longer execution time (max 10 hours). Auto-sklearn does not natively support text as input and we therefore execute it using universal sentence embedding representations omitting the time to extract those representations. We report the mean of 5 independent executions in terms of times and accuracy, noting little variance amongst the results.

\paragraph*{\underline{Baseline 3: Finetune}} The goal of this baseline is to replicate the SOTA values achieved for all datasets. We remark that this baseline is equipped with a strong prior knowledge which is usually unavailable for performing a cheap feasibility study and it only serves as a reference point. Unfortunately, reproducing the exact SOTA values was not possible for any of the dataset involved in the study, which is mainly due to computational constraints and the lack of publicly available reproducible code. We therefore perform our best, mostly manual, efforts to train a model on the original non-corrupted data. For multi-channel vision tasks (i.e., CIFAR10 and CIFAR100, and its noisy variants), we fine-tune EfficientNet-B4 using the proposed set of hyper-parameters~\cite{tan2019efficientnet}, whereas for NLP tasks, we fine-tune BERT-Base with 3 different learning rates and for 3 epochs~\cite{devlin2018bert}, using a maximal sequence length of 512, batch size of 6 and the Adam optimizer.

\subsection{Evaluation of BER Estimations}
\label{sec:endtoend}

We first evaluate \sysname by comparing its BER
estimation of the best achievable accuracy with
other baselines and show how this benefits an end-to-end 
scenario.

\paragraph*{\underline{\sysname vs. Baselines on Synthetic Noise}}

In Figure~\ref{fig:snoopy_vs_baselines} we present our main findings on three levels of realistically injected label noise --- 0\%, 20\% and 40\%, which we visualize by adding the increase of the SOTA (at the time of writing) in expectation as horizontal lines to indicate a proxy of the ground truth BER error. 
We see that \sysname is comparable to the short execution of auto-sklean whilst producing much better estimations. Furthermore, \sysname is much faster than all 
other methods, often by orders of magnitude.
The only exception is YELP in which running over large models (e.g., \textsc{GPT2} or \textsc{XLNet}) slows down \sysname in a fashion comparable to AutoKeras, whilst still outperforming it in terms of the estimated accuracy.
It also produces BER estimations
that are \textit{comparable, if not better}
than all other approaches.
In fact, it is often better than both LR and, particularly, AutoKeras. It is only slightly worse than the LR classifier on text tasks for IMDB and SST2,
while being orders of magnitude faster.

\paragraph*{\underline{\sysname vs. Baselines on Real Noise}}

In Figure~\ref{fig:snoopy_vs_real_baselines} we run the same set of experiments for real noisy variants of CIFAR10 and CIFAR100 from~\cite{wei2022learning}. We realize that \sysname constantly outperforms all baselines both in terms of speed and estimation accuracy. When comparing the error values to the lower and upper bounds, we realize that whilst \sysname remains inside the bounds, there is a considerable gap between them. Nonetheless, \sysname produces estimates close to the expected increase of the SOTA using Theorem~\ref{thm_flipping}.

\begin{figure}[t!]
\centering
\includegraphics[width=0.55\linewidth]{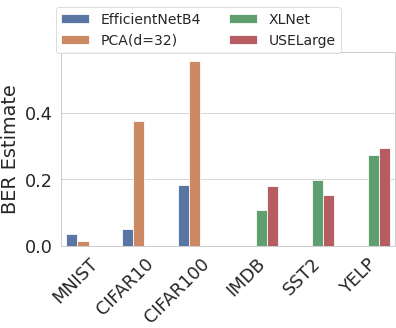}
\vspace{-1em}
\caption{Impact of fixing a single feature transformation.}
\vspace{-1em}
\label{fig:taking_min}
\end{figure}

\textit{\underline{Is Taking the Minimum Necessary?}} When analyzing the performance of the system with respect to the number of feature transformations, one might ask the question whether a single transformation always outperforms all the others and hence makes the selection of the minimal estimator obsolete. When conducting our experiments, we observed that selecting the \textit{wrong} embedding can lead towards a large gap when compared to the optimal embedding, e.g., favoring the embedding \textsc{USELarge} over \textsc{XLNet} on IMDB doubles the gap of the estimated BER to the known SOTA value~\cite{renggli2021evaluating}, whereas favoring \textsc{XLNet} over \textsc{USELarge} on SST2 increases the gap by 1.5$\times$, making proper selection necessary (c.f., Figure~\ref{fig:taking_min}).

\subsection{Usefulness of the Additional Guidance}
\label{secExpAdditionalGuidance}

\begin{figure}[t!]
\centering
\includegraphics[width=0.92\linewidth]{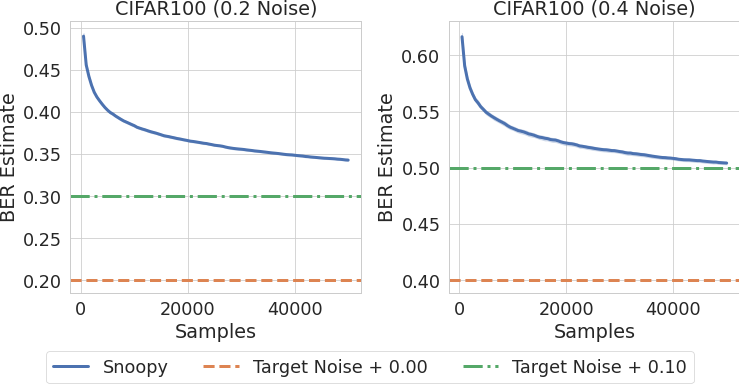}
\vspace{-1em}
\caption{CIFAR100 - Different noise levels and targets.}
\label{fig:additional_guidance_plots}
\vspace{-1em}
\end{figure}

\begin{figure}[t!]
\centering
\includegraphics[width=0.92\linewidth]{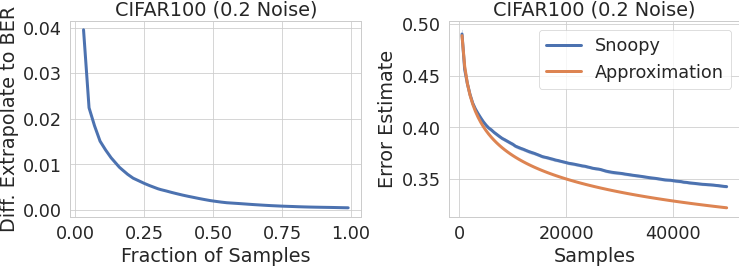}
\vspace{-1em}
\caption{CIFAR100 - (Left) Extrapolation accuracy to the full dataset based on a fraction of the samples. (Right) Approximation based on 5\% of the samples.}
\label{fig:additional_guidance_extrapolate}
\vspace{-1.5em}
\end{figure}

When evaluating the 1NN estimator accuracy for varying label noise, and its convergence under different feature transformations,
\footnote{We illustrate the convergence behaviour under different transformations in Appendix~\ref{app:Convergence} in the supplementary material.}
we see that even the best transformations are constantly over-estimating the lower bound when increasing label noise, validating the key arguments for taking the minimum over all estimators.
All the results indicate the median, 95\% and 5\% quantiles over multiple independent runs (i.e., 10 for YELP and 30 otherwise).
We observe the presence of much more instability in SST2 when compared to other datasets.
This is not at all surprising since SST2 has a very small test set consisting of less than one thousand samples, as seen in Table~\ref{tbl:datasets}. This naturally results in higher variance and less confidence for the 1NN classifier accuracy compared to the larger number of test samples for the other datasets.

Figure~\ref{fig:additional_guidance_plots} illustrates a convergence plot for a fixed embedding (EfficientNet-B5) and the clean CIFAR100 dataset injected with 20\% and 40\% label noise respectively. The two target accuracies visualized by a horizontal line represent exactly the noise level, and the noise level plus 10\%. Note that the noise level is only reachable if the original BER of the dataset is zero. From the visualizations, the target of 0.5 on the dataset with 0.4 noise is highly likely. By using the approximation from Equation~\ref{eqnApproxExtrap}, we realize that less than 10K more samples should suffice to attain this accuracy. Conversely, for a target of 0.3 and 0.2 noise, although possibly realizable, the number of additional samples to verify the quality of the extrapolation is already more than 260K. Note that Equation~\ref{eqnApproxExtrap} converges to zero, and therefore any target can be realizable. Targeting exactly the label noise for each of the datasets yields an extrapolated number of more than 16M and 84M, which both should be seen as not trustworthy approximations based on the much smaller number of available samples in the training set. This thus implies that the target accuracy is not achievable based on the given transformations and numbers of samples.
To illustrate this fact, we subsample the low label noise dataset and the same embedding for a fixed fraction. We then extrapolate the achievable target for the full dataset (i.e., 50K training samples) and plot the difference between the extrapolated target and the true BER estimate in Figure~\ref{fig:additional_guidance_extrapolate} on the left. The right part of Figure~\ref{fig:additional_guidance_extrapolate} illustrates the extrapolation based on 5\% of the samples. Notice that this provided example illustrates when to trust the estimated number of additional samples required using Equation~\ref{eqnApproxExtrap} (i.e., when the number if relatively low), not the BER estimate of \sysname. The same results can easily be shown for any other dataset.

\subsection{End-to-end Use Case}
\label{secEndToEnd}

\textit{How can we take advantage 
of \sysname to help practical use cases?} In this section,
we focus on a specific end-to-end use case of a feasibility study in which the user's task contains a target accuracy and a representative, but noisy dataset. The goal is to reach the target accuracy. 
At each step, the user can perform one of the following three actions: (1) clean a portion of the labels, (2) train a high-accuracy model using AutoKeras or fine-tune a state-of-the-art pre-trained model, (3) perform a feasibility study by either using the cheap LR model or \sysname. 
To simulate the cleaning process on a noisy dataset, which usually requires human interactions of an expert labeler, we focus on the manually polluted datasets with synthetic label noise, where we can simply \textit{restore} the original label from the dataset. Being aware of different human costs for cleaning labels in real-world scenarios (i.e., depending on the application and the required expertise), we compare the impact of different cost scenarios outlined below.
We report the mean (accuracy and run-time) over at least 5 independent runs.

\paragraph*{\underline{Different User Interaction Models}}

We differentiate two main scenarios in our end-to-end experimental evaluation: (1) \textit{without feasibility study} and (2) \textit{with feasibility study}. 
Without a feasibility study, users will start an expensive, high-accuracy run (i.e., running the fine-tuning baseline) using the input data. If the achieved accuracy is below the desired target, users will clean a fixed portion of the data (1\%, 5\%, 10\%, or 50\%, which we call \emph{steps}) and re-run the expensive training system. This is repeated until a model reaches the desired accuracy or all labels are cleaned. 
With a feasibility study, users alternate between running the feasibility study system and cleaning a portion of the data (set to 1\% of the data) until the feasibility study returns a positive signal or all labels are cleaned. Finally, a single expensive training run is performed. The lower bound on computation is given by training the expensive model exactly once.

\begin{figure}[t!]
\centering
\includegraphics[width=0.95\linewidth]{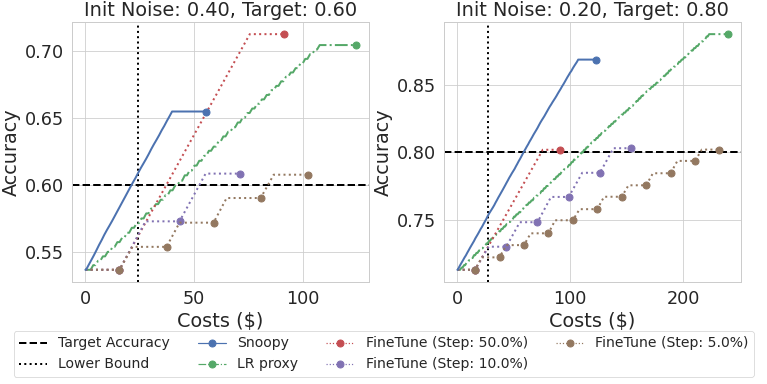}
\vspace{-1em}
\caption{CIFAR100 - End-to-end use case, cheap labels.}
\label{fig:end2end_cifar100_cheap}
\vspace{-1em}
\end{figure}

\begin{figure}[t!]
\centering
\includegraphics[width=0.95\linewidth]{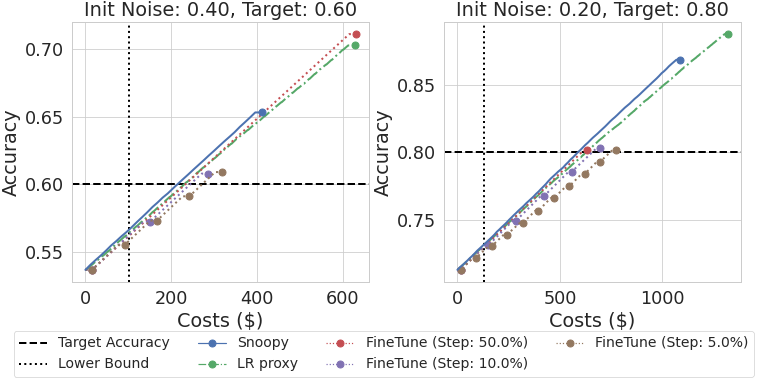}
\vspace{-1em}
\caption{CIFAR100 - End-to-end use case, expensive labels.}
\label{fig:end2end_cifar100_expensive}
\vspace{-1.5em}
\end{figure}

\paragraph*{\underline{Different Cost Scenarios}} We measure the cost in hypothetical ``dollar price'' for different regimes, depending on the \textit{human-labeling} costs and on the \textit{machine} costs. For the former, we define two scenarios: 'free', 'cheap' (0.002 dollars per label, resulting in 500 labels per dollar) and 'expensive' (0.02 dollars per label, resulting in 50 labels per dollar). For the latter, we fix the \textit{machine} cost to 0.9\$ per hour (the current cost of a single GPU Amazon EC2 instance).

\paragraph*{\underline{Key Findings}}
We only
present the results on CIFAR100 here and leave the rest to Appendix~\ref{secAppendixExtendedEval}
---we observe similar results on all datasets for a wide range of initial noise levels and target accuracies.
We show the results in Figures~\ref{fig:end2end_cifar100_cheap} and~\ref{fig:end2end_cifar100_expensive}, for 2 different cost setups described above (cheap and expensive), each over 2 values of the initial noise (0.40, 0.20) and, respectively, 2 target accuracies (0.60, 0.80). Each dot represents 
the result of one run of the expensive training process.

\paragraph*{(I) Feasibility Study Helps.} When comparing the costs of repetitively training an expensive model to those of using an efficient and accurate system that performs a feasibility study, such as \sysname, we see significant improvements across all results (c.f., blue vs. brown lines in Figure~\ref{fig:end2end_cifar100_cheap}). 
Without a system that performs a feasibility study, users are facing 
a dilemma. On the one hand, if one does not 
train an expensive model frequently 
enough, it might clean up 
more labels than necessary, to achieve the target accuracy, e.g., FineTune (step 50\%), which makes it intense on the human-labeling costs. This can be seen by the size of the vertical gap between the end point of a method and the horizontal line indicating the minimum number of samples to be cleaned to achieve the target accuracy. On the
other hand, if one trains an expensive model too frequently, visible in the number of stairs for expensive fine-tune lines or the steepness of the curves for faster methods,
it becomes computationally expensive, wasting a lot of
computation time. With a 
feasibility study, 
the user can balance these two factors better.
As running low-cost proxy
models is significantly 
cheaper than training an expensive model,
the user can get feedback 
more frequently (having in mind the efficient incremental implementation from Section~\ref{sectionEfficientImplementation}). Finally, notice that when we enter the label-cost dominated regime (e.g., Figure~\ref{fig:end2end_cifar100_expensive}), one seeks at cleaning the minimum amount of labels necessary, ignoring the computational costs. Nevertheless, finding the right step size is critical, making it a difficult task.

\paragraph*{(II) \sysname Outperforms Baselines.} 
When comparing different 
estimators that can be used
in a feasibility study, 
in most cases, 
\sysname is more effective 
compared to running a cheaper model such as LR, with its accuracy as a proxy. 
\sysname offers
significant savings compared to LR when the labeling costs are high.
The LR model will often be of a lower accuracy than 
an expensive approach; hence, it requires to clean more labels than necessary to reach the target.
We note that there are cases (e.g., for IMDB) where the best LR model yields a lower error than the BER estimator used by \sysname.
In such cases, there exists a regime where the costs of using the LR proxy are comparable or superior to using \sysname despite being more expensive to compute.
However, we see this as an exception and Figure~\ref{fig:end2end_cifar100_cheap}
(and the evaluations on other datasets in Appendix~\ref{secAppEnd2End})
clearly show that the LR proxy is usually significantly more costly than using \sysname.

\begin{figure}[t!]
\centering
\includegraphics[width=0.42\linewidth]{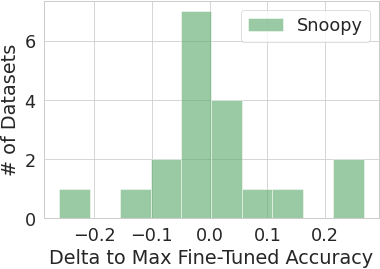}
\hfill
\includegraphics[width=0.42\linewidth]{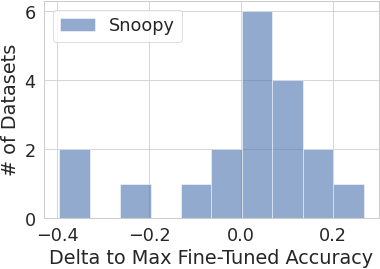}
\vspace{-0.5em}
\caption{\sysname vs. Maximal fine-tune accuracies on VTAB  (\textbf{Left}) for proprietary expert models derived from ~\cite{renggli2020model} and (\textbf{Right}) for public models from Huggingface: Showing \sysname's generalization ability on small datasets and embeddings trained on different tasks.}
\label{fig:vtab_results}
\vspace{-1em}
\end{figure}

\subsection{Generalization to Other Tasks}

In this section we examine two potential limitations of
\sysname, when deployed on a new task: (i) \textit{its dependence on large datasets}, and (ii) \textit{the necessity of having ``good'' pre-trained feature transformations for the given task}. For this, we use the results of~\cite{renggli2020model} on the popular visual task adaptation benchmark (VTAB)~\cite{zhai2019large} which is known to be a \textit{diverse} collection of datasets (19 different tasks), each being \textit{small} (1K training samples), and our collection of pre-trained transformations \textit{does not contain} any trained on these datasets. Additionally, we fine-tune the same 19 datasets on a set of 235 publicly available PyTorch models from Huggingface.

To validate that \sysname does not suffer from the above limitations, in Figure~\ref{fig:vtab_results} we illustrate the difference between \sysname's predictions and the best achieved post-fine-tune accuracies. We observe that on most datasets, \sysname produces a useful estimate of the fine-tune accuracy (except for some negative transfer results enabled by the low data regime) for both proprietary expert models and publicly available models.
The estimates of the later are slightly shifted to the right as expected. Even though this is sufficient to say that the currently available embeddings are supporting \sysname's performance, we expect this figure to improve over time as more and better embeddings become publicly available via repositories such as Huggingface, which also start including learned representation for different modalities such as tabular data.

\subsection{Efficiency of \sysname}
\label{secAccuracyAndEfficiency}

We saw that the gain of using \sysname comes from having an (i) \textit{efficient estimator} of (ii) \textit{high accuracy}. Those two requirements are naturally connected. While having access to more and ``better'' (pre-trained) transformations is key for getting a high accuracy of our estimator, it requires the implementation of our algorithm to scale with respect to the ever-increasing number of transformations. 

\begin{figure}
\centering
\includegraphics[width=0.93\linewidth]{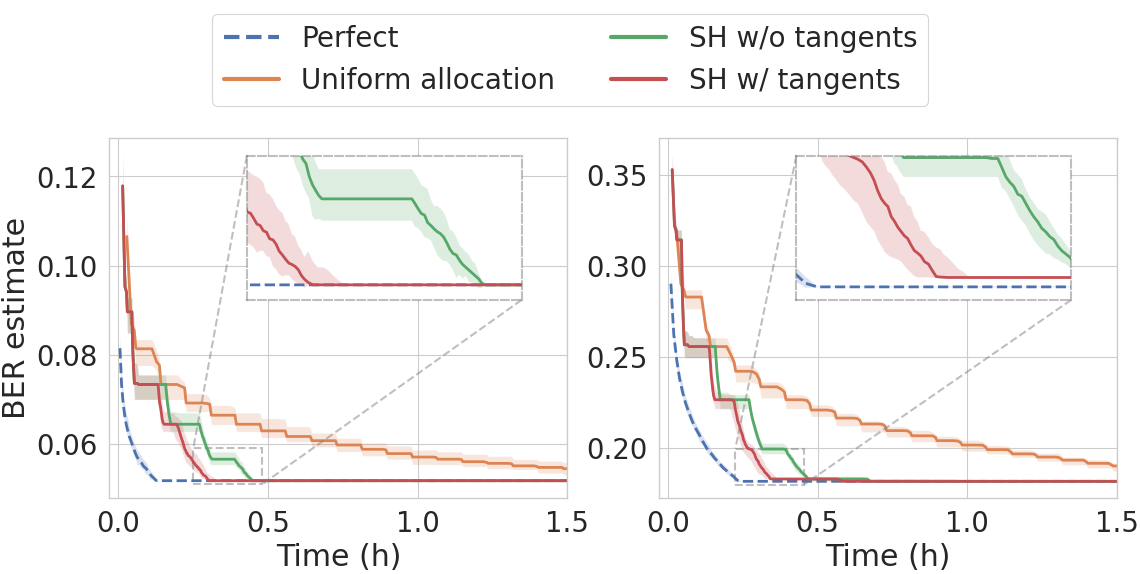}
\vspace{-1em}
\caption{Different selection strategies for \textbf{(Left)} CIFAR10, \textbf{(Right)} CIFAR100.}
\label{fig:strategy_cifar10_cifar100}
\vspace{-1em}
\end{figure}


\paragraph*{\underline{Runtime Analysis}} To showcase the importance of the \emph{successive-halving} (SH) algorithm, with and without the \emph{tangent method} presented in Section~\ref{sectionEfficientImplementation}, we compare different strategies for deploying the 1NN estimator in Figures~\ref{fig:strategy_cifar10_cifar100}.
The strategies are evaluated with respect to the runtime (averaged across multiple independent runs on a single Nvidia Titan Xp GPU) needed to reach an estimation within 1\% of the \textit{best} possible value using all the training samples.
Running the estimator only on the transformation yielding the minimal result is referred to as the \emph{perfect} strategy providing a lower bound, whereas we also test the \emph{uniform allocation} baseline described in~\cite{jamieson2016non}.  
We report the runtime by selecting the best batch size out of 1\%, 2\%, or 5\% of the training samples. We observe that running the entire feasibility study using \sysname on CIFAR100 on a single GPU takes slightly more than 16 minutes, whereas the largest examined NLP dataset YELP requires almost 8.5 hours, with a clear improvement of SH with the tangent method over the one without. Putting these numbers into context, fine-tuning EfficientNet-B4 on CIFAR100 on the same GPU with one set of hyper-parameters (out of the 56 suggested by the authors~\cite{tan2019efficientnet}) requires almost 10 hours (without knowing whether other embeddings would perform better), whereas training large NLP models usually requires several hundred accelerators~\cite{rasley2020deepspeed}.

\begin{figure}[t!]
\centering
\includegraphics[width=0.55\linewidth]{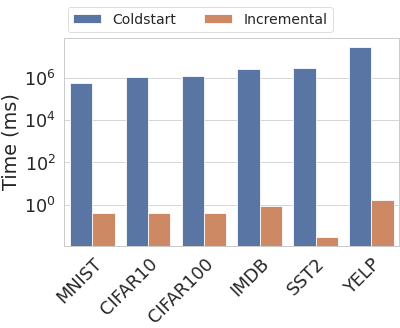}
\vspace{-1em}
\caption{Incremental execution (log scale): On all datasets, rerunning \sysname incrementally is several orders of magnitude faster than running from scratch.}
\label{fig:incremental}
\vspace{-1.5em}
\end{figure}

\paragraph*{\underline{Incremental Execution}} In Figure~\ref{fig:incremental}, not surprisingly, we see the benefit of \sysname's ability to quickly rerun incremental evaluation described in Section~\ref{sectionEfficientImplementation}. Compared to running from scratch, we see that incremental execution is faster by several orders of magnitude on all datasets.

\section{Conclusion}
\label{sectionConclusion}
\vspace{-0.5em}

We present \sysname, a novel system that enables a systematic feasibility study for ML application development. By consulting a range of estimators of the Bayes error and aggregating them in a theoretically justified way, \sysname suggests whether a predefined target accuracy is achievable. We demonstrate system optimizations that support the usability of \sysname, and scale with the increase in the number and diversity of available pre-trained embeddings in the future.

{
\small
\textbf{Acknowledgements.} 
CZ and the DS3Lab gratefully acknowledge the support from the Swiss State Secretariat for Education, Research and Innovation (SERI) under contract number MB22.00036 (for European Research Council (ERC) Starting Grant TRIDENT 101042665), the Swiss National Science Foundation (Project Number 200021\_184628, and 197485), Innosuisse/SNF BRIDGE Discovery (Project Number 40B2-0\_187132), European Union Horizon 2020 Research and Innovation Programme (DAPHNE, 957407), Botnar Research Centre for Child Health, Swiss Data Science Center, Alibaba, Cisco, eBay, Google Focused Research Awards, Kuaishou Inc., Oracle Labs, Zurich Insurance, and the Department of Computer Science at ETH Zurich.
}

\bibliographystyle{IEEEtran}
\bibliography{IEEEabrv,references}

\onecolumn

\section{Class-Dependent Label Noise}\label{app:label_flipping}
In this Section we provide the proof of Theorem~\ref{thm_flipping} and provide further discussion on the main result. 

Recall that we suppose that the noise in the labels is class dependent, rather than instance dependent. In other words, for a noisy random variable $\tY$, we assume
\[
t_{\ty,y} := \PP (Y_\rho = \ty \ | Y = y, X = x) = \PP (Y_\rho = \ty \ | Y = y)
\]
corresponding to the transition matrix $T$ in~\cite{wei2022learning}. Moreover, we assume $y_x = \argmax_{y\in \cY} p(Y=y | x) = \argmax_{y \in \cY} p_\rho (Y_\rho =y | x)$.

\begin{proofof}{Theorem~\ref{thm_flipping}}
We have
\begin{align*}
    R_{X, Y_\rho}^* &= \E_{X} [ 1 - \PP (Y_\rho = y_x | x) ] \\
    &= 1 - \E_{X} \left[ \sum_{y\in \cY} \PP (Y_\rho = y_x, Y = y | X = x) \right] \\
    &= 1 - \E_{X} \left[ \sum_{y\in \cY} \PP (Y_\rho = y_x | Y = y, X = x) \PP (Y = y | X = x) \right] \\
    &= 1 - \E_{X} \left[ \sum_{y\in \cY} t_{y_x, y} p (y | x) \right] \\
    &= 1 -  \E_{X} [t_{y_x, y_x} p(y_x | x)] - \E_{X} \left[ \sum_{y \neq y_x} t_{y_x, y} p (y | x) \right] \\
    &= \E_X [1 - p(y_x | x) ] + \E_X [\rho (y_x) p(y_x | x)] - \E_{X} \left[ \sum_{y \neq y_x} t_{y_x, y} p (y | x) \right],
\end{align*}
recalling that $\rho (y) = 1-t_{y,y}$ is the flipping fraction with respect to $y$.

This implies that 
\begin{equation}
    \label{eqn_noise_first}
    R_{X,Y_\rho}^* = R_{X,Y}^* + \E_X [\rho (y_x) p(y_x | x)] - \E_{X} \left[ \sum_{y \neq y_x} t_{y_x, y} p (y | x) \right],
\end{equation}
concluding the proof.
\end{proofof}

\paragraph{Getting lower and upper bounds on the BER with respect to $Y_\rho$}

To get lower and upper bounds from~(\ref{eqn_noise_first}), note that 
\begin{equation}\label{eqn_noise_third}
\min_{y} \rho (y) \leq \rho (y_x) \leq \max_{y} \rho (y) \hspace{3mm}\Longleftrightarrow \hspace{3mm} \min_{y} (1-t_{y,y}) \leq \rho (y_x) \leq \max_{y} (1-t_{y,y}),
\end{equation}
and, for $y \neq y_x$,
\begin{equation}\label{eqn_noise_fourth}
-\max_{y',y'': y'\neq y''} t_{y',y''} \leq -t_{y_x, y} \leq -\min_{y',y'': y'\neq y''} t_{y',y''}.
\end{equation}
Comparing with Figure 3 in~\cite{wei2022learning}, (\ref{eqn_noise_third}) corresponds to the smallest and largest distance from $1$ on the diagonal, and (\ref{eqn_noise_fourth}) corresponds to the smallest and largest non-diagonal elements.

Plugging in the two lower bounds of (\ref{eqn_noise_third}) and (\ref{eqn_noise_fourth}) into (\ref{eqn_noise_first}), we get
\begin{align}\label{eqn_noise_lb}
    R_{X,Y_\rho}^* &\geq R_{X,Y}^* + \min_{y}(1-t_{y,y}) \E_{X} [p(y_x | x)] - \max_{y',y'' \colon y'\neq y''} t_{y',y''} \E_{X} \left[ \sum_{y\neq y_x} p(y|x)\right] \nonumber \\
    &= R_{X,Y}^* + \min_{y}(1-t_{y,y}) \E_{X} [p(y_x | x)] - \max_{y',y'' \colon y'\neq y''} t_{y',y''} \E_{X} \left[ 1-p(y_x|x)\right] \\
    &= R_{X,Y}^* + (1-R_{X,Y}^*)\min_{y}(1-t_{y,y}) - R_{X,Y}^* \max_{y',y'' \colon y'\neq y''} t_{y',y''}. 
\end{align}
Plugging in the two upper bounds yields
\begin{align}\label{eqn_noise_ub}
    R_{X,Y_\rho}^* &\leq R_{X,Y}^* + (1-R_{X,Y}^*)\max_{y\in \cY}(1-t_{y,y}) - R_{X,Y}^* \min_{y',y'' \colon y'\neq y''} t_{y',y''}. 
\end{align}
Merging (\ref{eqn_noise_lb}) and (\ref{eqn_noise_ub}) yields 
\begin{align}\label{eqn_noise_lb_and_ub}
    R_{X,Y_\rho}^* \in &\left[ R_{X,Y}^* + (1-R_{X,Y}^*) \min_{y\in \cY}(1-t_{y,y}) - R_{X,Y}^* \max_{y',y'' \colon y'\neq y''} t_{y',y''} , \right. \\ &\left. \hspace{3mm} R_{X,Y}^* + (1-R_{X,Y}^*) \max_{y\in \cY}(1-t_{y,y}) - R_{X,Y}^* \min_{y',y'' \colon y'\neq y''} t_{y',y''} \right].
\end{align}
Using $0\leq R_{X,Y}^* \leq s_{X,Y}$, where $s_{X,Y}$ denotes the error of state-of-the-art model, we get
\begin{equation}\label{eqn_noise_final}
    R_{X,Y_\rho}^* \in \left[ (1-s_{X,Y}) \min_{y\in \cY}(1-t_{y,y}) - s_{X,Y} \max_{y',y'' \colon y'\neq y''} t_{y',y''} , \hspace{3mm} s_{X,Y} + \max_{y\in \cY}(1-t_{y,y}) \right].
\end{equation}
These represent valid lower and upper bounds, whereas in experimental section we also plot the following approximation of the RHS of (\ref{eqn_noise_first}):
\begin{equation}\label{eqn_predicted_mean}
R_{X,Y_{\overline{\rho}}} = s_{X,Y} + \E_{Y} \rho (y) (1-s_{X,Y}),
\end{equation}
representing average distance from 1 of diagonal elements, instead of simply taking minimum and maximum of the distances.

\paragraph{Examples.} We now provide two examples of how one could use Theorem~\ref{thm_flipping}, noting that the first example yields Lemma~\ref{lemLabelFlipping}.
\begin{itemize}
    \item \textit{(Uniform flipping)} In this scenario we assume that for each class we flip $\rho$ fraction of examples to any other class (including the original one) uniformly at random. In other words, 
    \[
        \rho(y) = \rho \cdot \left( 1 - \frac{1}{C} \right), \forall y, \hspace{5mm} t_{y',y''} = \frac{\rho}{C}, \forall y'\neq y''. 
    \]
    In this case Equation~\ref{eqn_noise_first} becomes 
    \begin{align*}
        R_{X,Y_\rho}^* &= R_{X,Y}^* + \rho \left( 1-\frac{1}{C} \right) \E_{X} [p(y_x | x)] - \frac{\rho}{C} \E_{X} \left[ \sum_{y\neq y_x} p(y|x) \right] \\ 
        &=  R_{X,Y}^* + \rho \left( 1-\frac{1}{C} \right) \E_{X} [p(y_x | x)] - \frac{\rho}{C} \left(1-\E_{X} \left[ p(y_x|x) \right] \right) \\
        &= R_{X,Y}^* + \rho \E_{X} [p(y_x | x)] - \frac{\rho}{C} \\
        &= R_{X,Y}^* + \rho \left( 1 - \frac{1}{C} - R_{X,Y}^*\right).
    \end{align*}
    We see that this is exactly Lemma~3.1 in~\cite{renggli2021evaluating}.
    
    \item \textit{(Pairwise flipping)} In this scenario we assume that for each $y$ there exists a single class $y_x'$ such that $p(y_x'|x) = 1 - p(y_x|x)$. In that case, we flip $\rho$ fraction of samples to $y_x'$. That means that $\rho (y_x) = 1- t_{y_x, y_x'} = \rho $, and all other $t_{y_x,y} = 0$, for $y\neq y_x', y_x$. Then Equation~\ref{eqn_noise_first} becomes
    \begin{align*}
    R_{X,Y_\rho}^* &= R_{X,Y}^* + \rho \E_X [ p(y_x | x)] - \rho \E_{X} \left[ p (y_x' | x) \right] \\
    &= R_{X,Y}^* + \rho \left( 2\E_X [p(y_x | x)] - 1 \right) \\
    &= R_{X,Y}^* + \rho \left( 1-2R_{X,Y}^* \right).
    \end{align*}
    
\end{itemize}

\paragraph{Bounds for CIFAR-N} Inserting the values from Figure 3 in~\cite{wei2022learning} into Equation~\ref{eqn_noise_final} yields:
    \begin{align*}
        \text{CIFAR10-N Aggregated } &: \hspace{3mm} R_{X,Y_\rho} \in \left[ 0.03 - 0.10 \cdot s_{X,Y}, \hspace{3mm} 0.17 + s_{X,Y} \right] \\
        \text{CIFAR10-N Random } &: \hspace{3mm} R_{X,Y_\rho} \in  \left[ 0.10 -0.23 \cdot s_{X,Y} , \hspace{3mm} 0.26 + s_{X,Y} \right] \\
        \text{CIFAR100-N Noisy } &: \hspace{3mm}  R_{X,Y_\rho} \in \left[ 0.083 - 0.312 \cdot s_{X,Y} , \hspace{3mm} 0.854 + s_{X,Y} \right].
    \end{align*}

\clearpage
\newpage

\section{Justifications on Taking the Minimum}
\label{app:justifications_min}

\begin{figure}[ht!]
\centering
\includegraphics[width=0.6\linewidth]{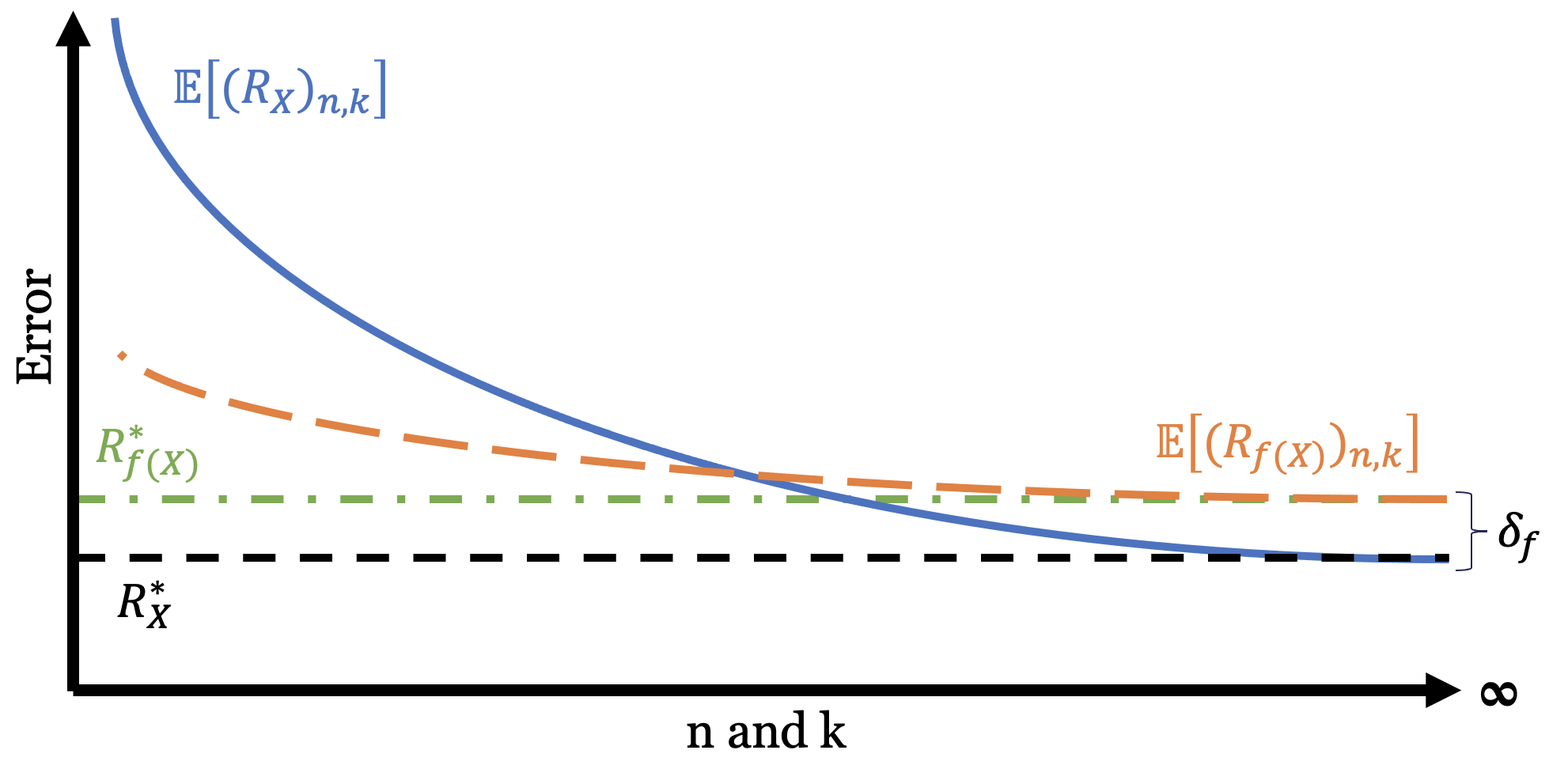}
\caption{Transformation bias $\delta_f$ for a fixed transformation $f$.}
\label{fig:illustration_transformation_bias}
\end{figure}

\begin{figure}[ht!]
\centering
\includegraphics[width=0.6\linewidth]{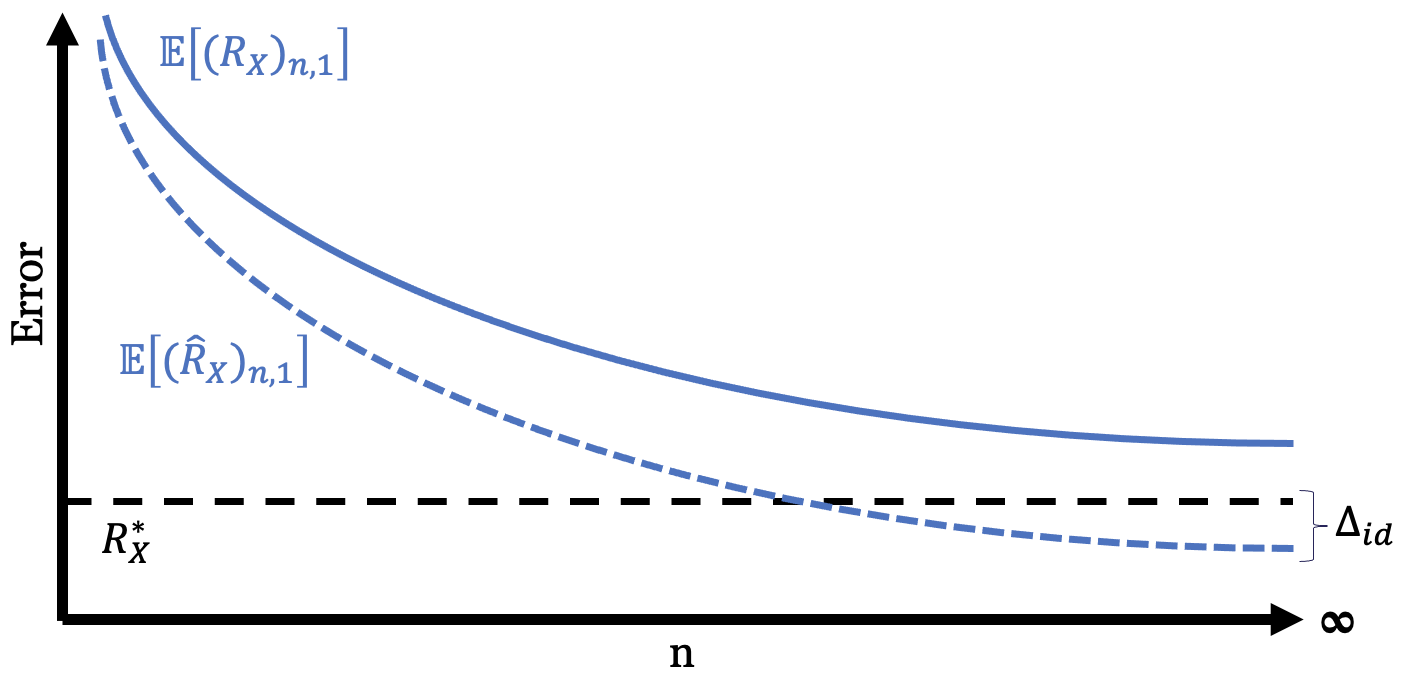}
\caption{Asymptotic tightness $\Delta_{id}$ on the raw data (identify transformation).}
\label{fig:illustration_asymptotic_tighness_id}
\end{figure}

\begin{figure}[ht!]
\centering
\includegraphics[width=0.6\linewidth]{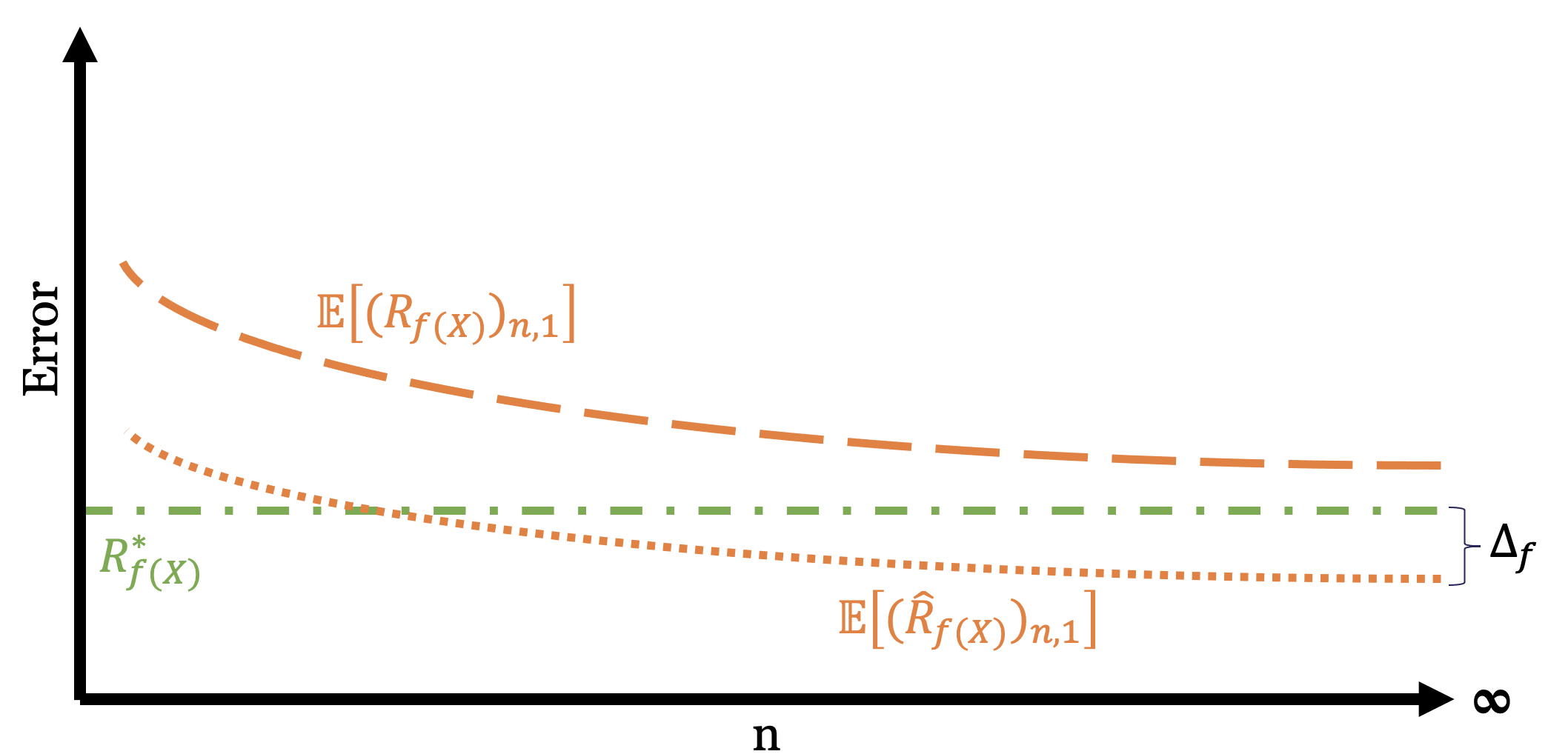}
\caption{Asymptotic tightness $\Delta_f$ for a fixed transformation $f$.}
\label{fig:illustration_asymptotic_tighness_f}
\end{figure}

\begin{figure}[ht!]
\centering
\includegraphics[width=0.6\linewidth]{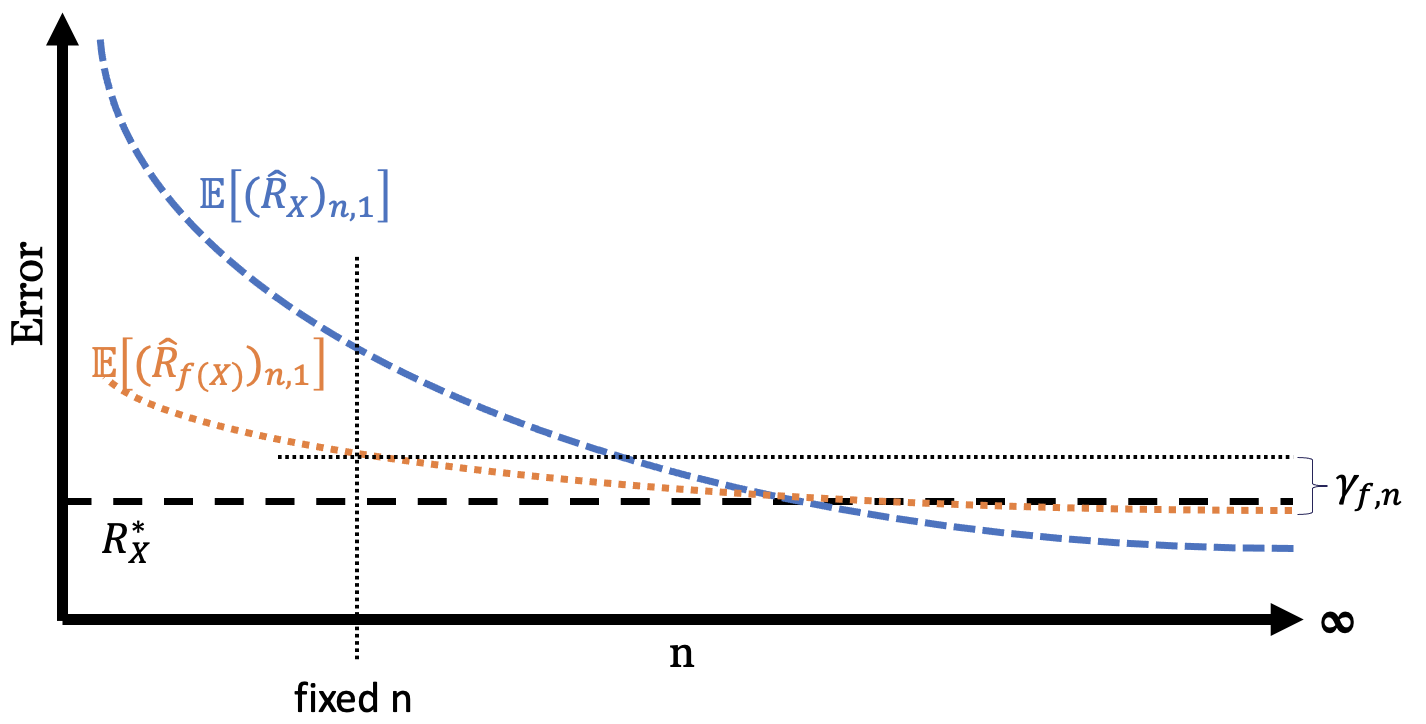}
\caption{N-sample gap (of the estimator) $\gamma_{f,n}$ and a single transformation $f$.}
\label{fig:illustration_finite_sample_bias}
\end{figure}

Figures~\ref{fig:illustration_transformation_bias}-~\ref{fig:illustration_finite_sample_bias} illustrate the quantities defined in Section~\ref{sectionTheoreticalAnalysis}.

\FloatBarrier

\FloatBarrier

\subsection{Proof of Sufficient Conditions}

We next proof that the conditions of Cases 1 and 2 are sufficient for the claims maid in the main body of this work.

\paragraph{Case 1: When is $\RH$ optimal?}

We prove that Condition~\ref{eqnCase1} implies $\widehat{R}_{f(X),n}\geq R_{X}^*$, for all $f\in\cF$.

\begin{proofcap}
For any transformation $f$,

\begin{align*}
    \widehat{R}_{f(X),n} &= \widehat{R}_{f(X),\infty} + \gamma_{f,n} \\
    &= R_{f(X)}^* - \Delta_f + \gamma_{f,n} \\
    &= R_{X}^* + \delta_f + \gamma_{f,n} - \Delta_f \\
    &\geq R_{X}^*,
\end{align*}
with the last inequality coming from Condition~\ref{eqnCase1}.
\end{proofcap}

\paragraph{Case 2: When is $\RH$ at least as good as the suggested estimator by \cite{Cover1967-zg} on the raw features?}

We prove Condition~\ref{eqnCase2} implies $\widehat{R}_{f(X),n}\geq \widehat{R}_{X,\infty}$, for all $f\in \cF$.

\begin{proofcap}
For any transformation $f$,

\begin{align*}
    \widehat{R}_{f(X),n} &= \widehat{R}_{f(X),\infty} + \gamma_{f,n} \\
    &= R_{X}^* + \delta_f + \gamma_{f,n} - \Delta_f \\
    &\geq R_{X}^* - \Delta_{id} \\
    &=\widehat{R}_{X,\infty},
    \end{align*}
    where the inequality comes from Condition~\ref{eqnCase2}.
\end{proofcap}

\clearpage
\section{Successive-Halving with Tangents}
\label{app_SH_tangents}

In Section~\ref{sectionEfficientImplementation} we illustrated the successive-halving algorithm together with the improvement that uses tangent predictions to avoid unnecessary calculations on transformations that will certainly not proceed to the next step. In this section we provide an illustration of the algorithm in Figure~\ref{fig:improved_SH} along with pseudocodes for both variants of the successive-halving algorithm. Switching from one variant to the other is simply done through the \verb|use_tangent| flag, which either calls the function \verb|Pulls_with_tangent_breaks|, or avoids this and performs the original successive-halving algorithm. As the remaining transformations after each step are the same in both variants, all theoretical guarantees of successive-halving can be transferred to our extension.

The \verb|Pulls_with_tangent_breaks| function simply uses the tangent, which we approximate by a line through the two last known points of the convergence curve, to predict the smallest error that a feature transformation can achieve at the end of the current step. Here we assume that the convergence curves are convex, which holds on average~\cite{snapp1991asymptotic}. Therefore, it is a slightly more aggressive variant of the successive-halving algorithm, but we did not observe failures in practice since the tangent is usually a very crude lower bound. 

\FloatBarrier

\begin{algorithm}[ht]
   \caption{Successive-Halving with Tangents}
   \label{alg:SHwT}
\begin{algorithmic}
   \STATE {\bfseries Input:} Flag use\_tangent, budget $B$, arms $1, \ldots, n$  with $l_{i,k}$ denoting the $k$-th loss from the $i$-th arm \\
    \STATE {\bfseries Initialize:} $S_0 = [ n ]$, predictions[$i$] = 0, for all $i\in [n]$%
    \FOR{$k=0,1,\ldots, \lceil \log_2 (n)\rceil - 1$}
    \STATE $L=|S_k|$;
    \STATE $r_k = \lfloor \frac{B}{L\cdot \lceil \log_2 (n) \rceil} \rfloor$
    \STATE $R_k = \sum_{j=0}^{k} r_j$
    \STATE Pull $r_k$ times each arm $i = 1,\ldots, \lfloor L/2 \rfloor$ 
        \IF{{\bfseries not} use\_tangent}
             \STATE Pull $r_k$ times each arm $i = \lfloor L/2 \rfloor + 1, \ldots, L$    
        \ELSE
            \STATE threshold $= \max_{i=1,\ldots, \lfloor L/2 \rfloor} l_{i,R_k}$
            \STATE $S_k$ = Pulls\_with\_tangent\_breaks($S_k$, $r_k$, $R_k$, predictions[], threshold)
        \ENDIF
        \STATE Let $\sigma_k$ be a permutation on $S_k$ such that $l_{\sigma_k(1), R_k} \leq \ldots \leq l_{\sigma_k(|S_k|), R_k}$
    \STATE $S_{k+1} = \{ \sigma_k(1), \ldots, \sigma_k(\lfloor L/2 \rfloor)\}$.
    \ENDFOR
    \STATE {\bfseries Output:} Singleton element of $S_{\lceil \log_2 (n) \rceil}$ \\    
\end{algorithmic}
\end{algorithm}

\begin{algorithm}[ht]
   \caption{Pulls\_with\_tangent\_breaks}
\begin{algorithmic}
   \STATE {\bfseries Input:} $S$, $r$, $R$, predictions[], threshold
   \FOR{$i \in S$}
    \FOR{$j=1,\ldots, r$}
    \IF{predictions[i]$>$threshold}
        \STATE remove $i$ from $S$
        \STATE {\bfseries break}
    \ELSE 
    \STATE Pull arm $i$ once
    \STATE Update predictions[$i$] using tangent value at $R$
    \ENDIF
    \ENDFOR
    \ENDFOR
    \STATE {\bfseries Return} $S$ 
\end{algorithmic}
\end{algorithm}


\clearpage
\section{Extended Experimental Evaluation}
\label{secAppendixExtendedEval}

\subsection{Feature Transformations}
\label{secAppendixTransormations}

We report all the tested feature transformations for \sysname-enabled estimators in Table~\ref{tbl:tested_embeddings_images} for computer vision tasks, and in Table~\ref{tbl:tested_embeddings_nlp} for NLP tasks. All pre-trained embeddings can be found using the transformation name and the corresponding source\footnote{TensorFlow Hub: \url{https://tfhub.dev}, PyTorch Hub: \url{https://pytorch.org/hub}, HuggingFace Transformers: \url{https://huggingface.co/transformers/} and scikit-learn: \url{https://scikit-learn.org/}}.
In case of the HuggingFace Transformers embeddings in Table~\ref{tbl:tested_embeddings_nlp}, the \textit{Pooled} version denotes the representation that the embedding outputs for the entire text.
For other embeddings from that library, the representation of the text is computed by taking the mean over the representations of individual input tokens.
PCA transformations are trained on the training set and applied to both the training and the test set.
The inference on the training and the test set is run by batching multiple data points as shown in the last column of the tables.
The inference sizes are chosen such that the hardware accelerator is capable of running the inference without running out of memory for each transformation when executed in parallel with the full system. There is no limitation when using the identity transformation (raw), or running PCA, which is executed without any accelerator.
In the image domain, pre-trained embeddings assume a fixed-size resolution that might differ from the target image size. We adjust those cases using default resizing methods of either TensorFlow or PyTorch. The \emph{identity} transformation allows us to quantify the difference between the raw data representation and transformed features, albeit only for the vision tasks.

\FloatBarrier

\begin{table}[ht]
\centering
\caption{Feature transformations for vision tasks.}
\label{tbl:tested_embeddings_images}
\begin{tabular}{@{}lcllc@{}}
\toprule
Transformation                         & $\textsc{Dim}$  & Source   & Infer. Size \\ \midrule
\emph{Identity (Raw)} & -             & -         & N/A        \\
$\rm{PCA}_{32}$                        & 32   & scikit-learn           & N/A        \\
$\rm{PCA}_{64}$                        & 64   & scikit-learn           & N/A        \\
$\rm{PCA}_{128}$                       & 128  & scikit-learn           & N/A        \\
AlexNet                                & 4096 & PyTorch Hub            & 500        \\
GoogLeNet                              & 1024 & PyTorch Hub            & 500        \\
VGG16                                  & 4096 & PyTorch Hub            & 125        \\
VGG19                                  & 4096 & PyTorch Hub            & 125        \\
InceptionV3                            & 2048 & TensorFlow Hub         & 250        \\
ResNet50-V2                            & 2048 & TensorFlow Hub         & 250        \\
ResNet101-V2                           & 2048 & TensorFlow Hub         & 250        \\
ResNet152-V2                           & 2048 & TensorFlow Hub         & 250        \\
EfficientNet-B0                        & 1280 & TensorFlow Hub         & 125        \\
EfficientNet-B1                        & 1280 & TensorFlow Hub         & 125        \\
EfficientNet-B2                        & 1408 & TensorFlow Hub         & 125        \\
EfficientNet-B3                        & 1536 & TensorFlow Hub         & 125        \\
EfficientNet-B4                        & 1792 & TensorFlow Hub         & 50         \\
EfficientNet-B5                        & 2048 & TensorFlow Hub         & 25         \\
EfficientNet-B6                        & 2304 & TensorFlow Hub         & 20         \\
EfficientNet-B7                        & 2560 & TensorFlow Hub         & 10         \\ \bottomrule
\end{tabular}
\end{table}

\begin{table}[th]
\centering
\caption{Feature transformations for NLP tasks.}
\label{tbl:tested_embeddings_nlp}
\begin{tabular}{@{}lcllc@{}}
\toprule
Transformation                   & $\textsc{Dim}$ & Source                    & Infer. Size \\ \midrule
NNLM-en                          & 50             & TensorFlow Hub          & 5000       \\
NNLM-en (With Normalization)     & 50             & TensorFlow Hub          & 5000       \\
NNLM-en                          & 128            & TensorFlow Hub         & 5000       \\
NNLM-en (With Normalization)     & 128            & TensorFlow Hub          & 5000       \\
ELMo                             & 1024           & TensorFlow Hub              & 4          \\
Universal Sentence Encoder (USE)       & 512            & TensorFlow Hub            & 2500       \\
Universal Sentence Encoder (USE) Large & 512            & TensorFlow Hub           & 50         \\
BERT Base Cased Pooled (BCP)           & 768            & HuggingFace Transformers & 100        \\
BERT Base Uncased Pooled (BUP)         & 768            & HuggingFace Transformers & 100        \\
BERT Base Cased (BC)           & 768            & HuggingFace Transformers & 100        \\
BERT Base Uncased (BU)         & 768            & HuggingFace Transformers & 100        \\
BERT Large Cased Pooled (LCP)          & 1024           & HuggingFace Transformers & 50         \\
BERT Large Uncased Pooled (LUP)        & 1024           & HuggingFace Transformers & 50         \\
BERT Large Cased (LC)         & 1024           & HuggingFace Transformers & 50         \\
BERT Large Uncased (LU)       & 1024           & HuggingFace Transformers & 50         \\
XLNet                     & 768            & HuggingFace Transformers & 25         \\
XLNet Large              & 1024           & HuggingFace Transformers & 25         \\ \bottomrule
\end{tabular}
\end{table}


\clearpage
\subsection{Convergence Plots}\label{app:Convergence}

\begin{figure*}[ht]
\centering
\subfigure[MNIST]{\includegraphics[width=0.48\textwidth]{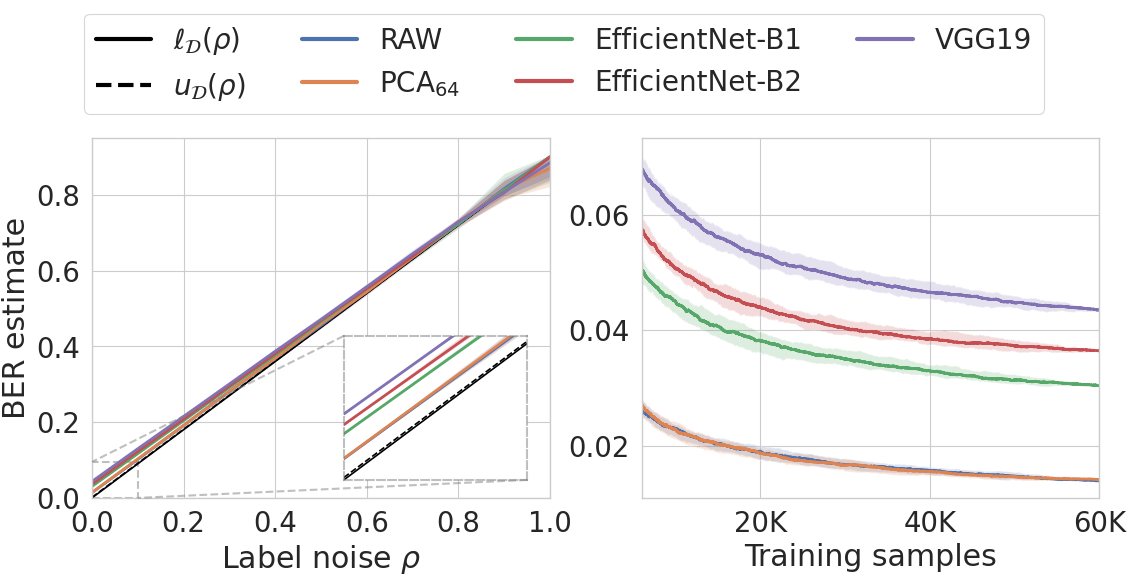}}
\subfigure[CIFAR10]{\includegraphics[width=0.48\textwidth]{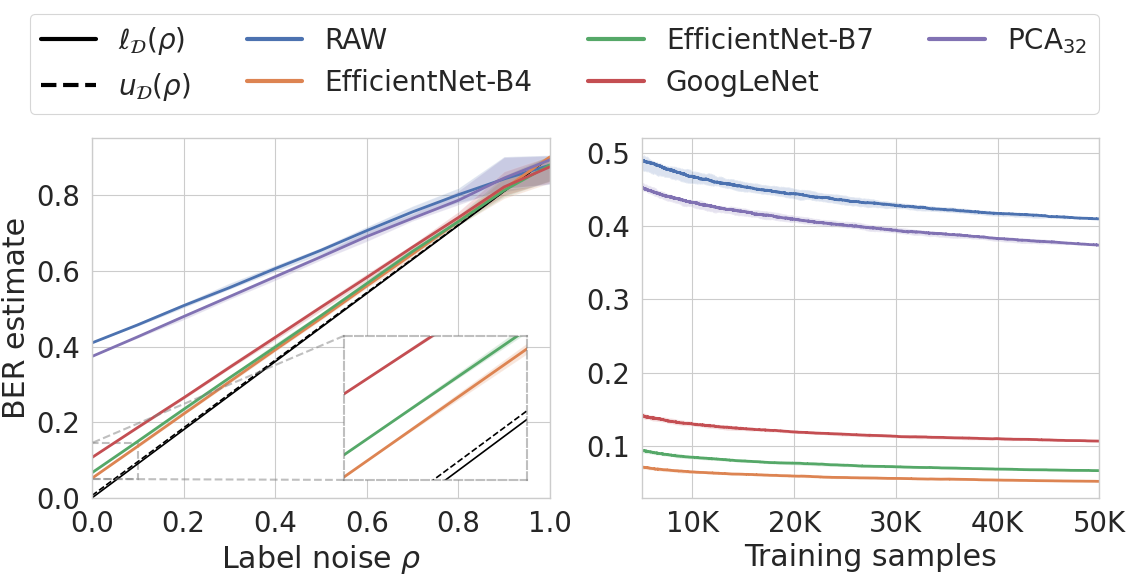}}
\caption{Evaluation and convergence of 1NN estimator for different feature transformations. \textbf{(Left each)} All the training points for different amount of label noise. \textbf{(Right each)} Zero label noise and increasing number of training samples.}
\label{fig:1nn_mnist_cifar10}
\end{figure*}

\begin{figure*}[ht]
\centering
\subfigure[IMDB]{\includegraphics[width=0.48\textwidth]{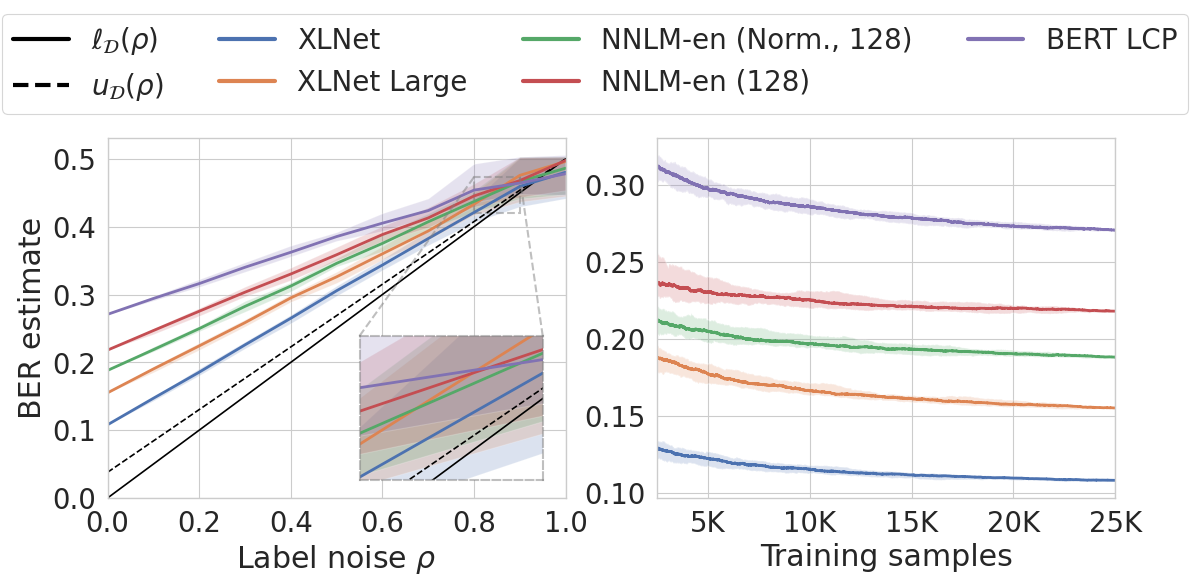}}
\subfigure[SST2]{\includegraphics[width=0.48\textwidth]{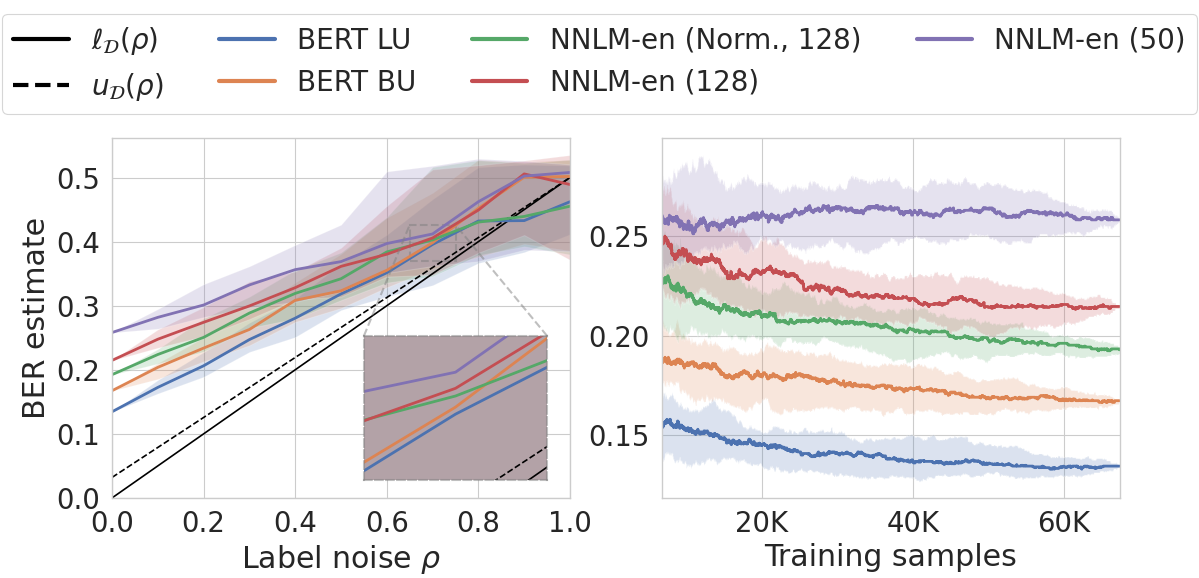}}
\caption{Evaluation and convergence of 1NN estimator for different feature transformations. \textbf{(Left each)} All the training points for different amount of label noise. \textbf{(Right each)} Zero label noise and increasing number of training samples.}
\label{fig:1nn_imdb_sst2}
\end{figure*}

\begin{figure*}[ht]
\centering
\subfigure[CIFAR100]{\includegraphics[width=0.45\textwidth]{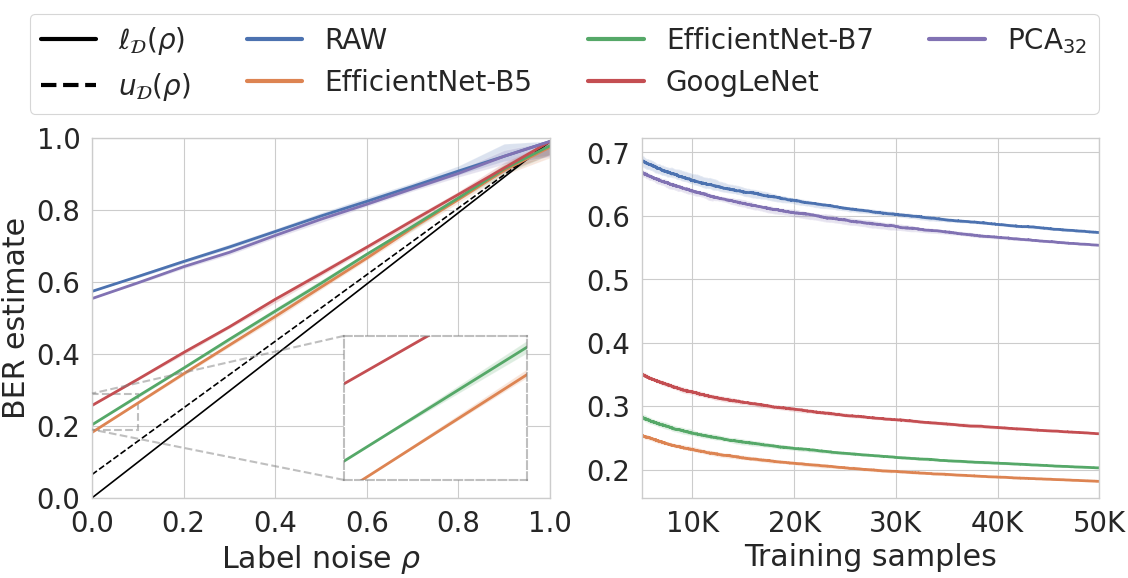}}
\subfigure[YELP]{\includegraphics[width=0.45\textwidth]{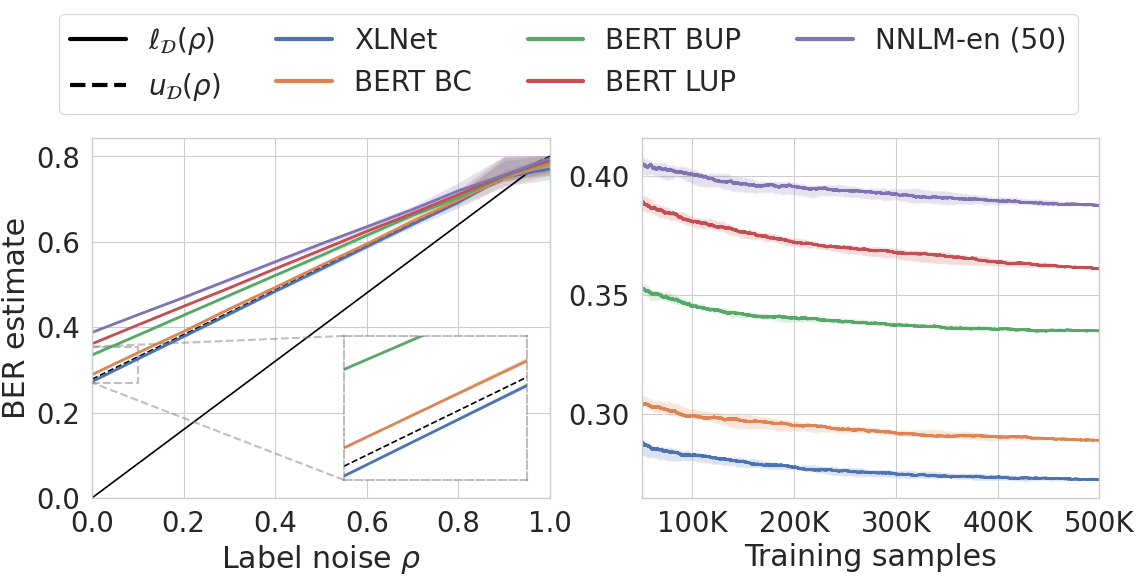}}
\caption{Evaluation and convergence of 1NN estimator for different feature transformations. \textbf{(Left each)} All the training points for different amount of label noise. \textbf{(Right each)} Zero label noise and increasing number of training samples.}
\label{fig:1nn_cifar100_yelp}
\end{figure*}

We provide the convergence plots and the evolution of the BER for all datasets and various embeddings (including the best performing one) in Figures~\ref{fig:1nn_mnist_cifar10}-\ref{fig:1nn_cifar100_yelp}.

\FloatBarrier

\FloatBarrier

\subsection{Additional End-To-End Figures}
\label{secAppEnd2End}

We report the figures for the end-to-end use-case for the omitted datasets next. We did not run the experiments for MNIST, where fine-tuning a large model performed worse than running the AutoML system. Nonetheless, following the insight of Section~\ref{secEndToEnd} and Figure~\ref{fig:snoopy_vs_baselines}, we expect no changes in the insights.

\FloatBarrier

\begin{figure}[ht]
\centering
\includegraphics[width=0.5\linewidth]{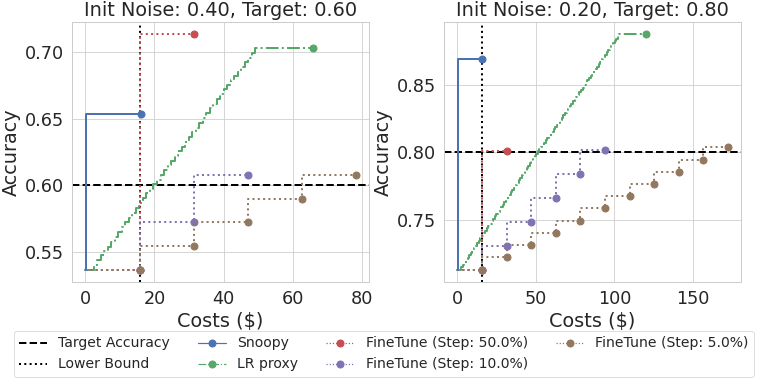}
\caption{CIFAR100 - End-to-end use case, free labels.}
\label{fig:end2end_cifar100_free}
\end{figure}

\FloatBarrier

\begin{figure*}[ht]
\centering
\subfigure[CIFAR10 - Free labels.]{\includegraphics[width=0.45\textwidth]{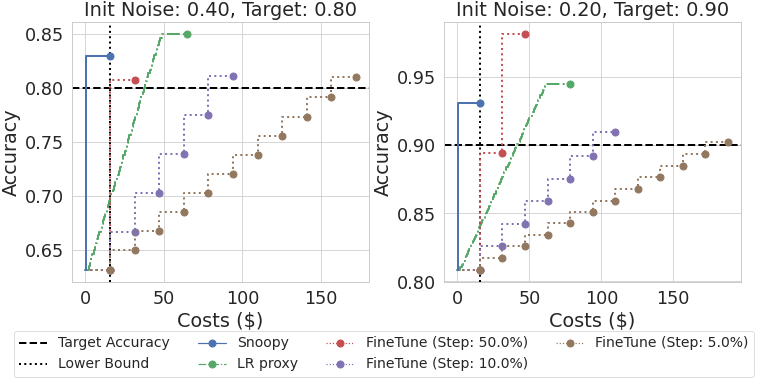}}
\subfigure[CIFAR10 - Cheap labels.]{\includegraphics[width=0.45\textwidth]{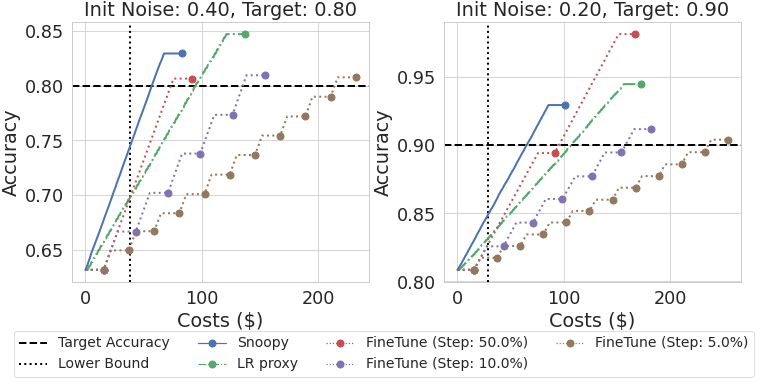}}
\caption{End-to-end use case: (a) CIFAR10 - free labels, and (b) CIFAR10 - cheap labels.}
\label{fig:end2end_cifar10_free_cifar10_cheap}
\end{figure*}

\begin{figure*}[ht]
\centering
\subfigure[CIFAR10 - Expensive labels.]{\includegraphics[width=0.45\textwidth]{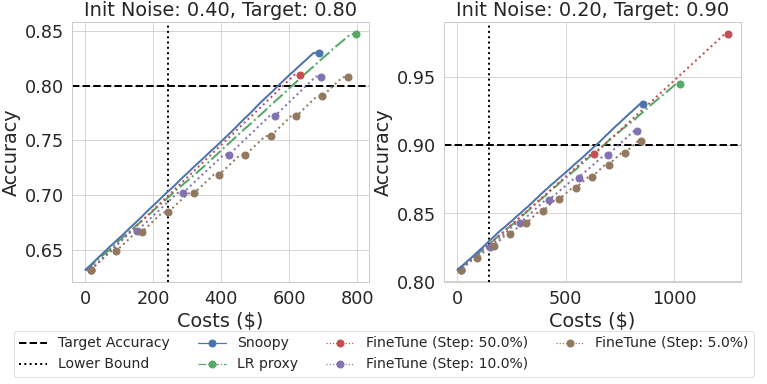}}
\subfigure[IMDB - Free labels.]{\includegraphics[width=0.45\textwidth]{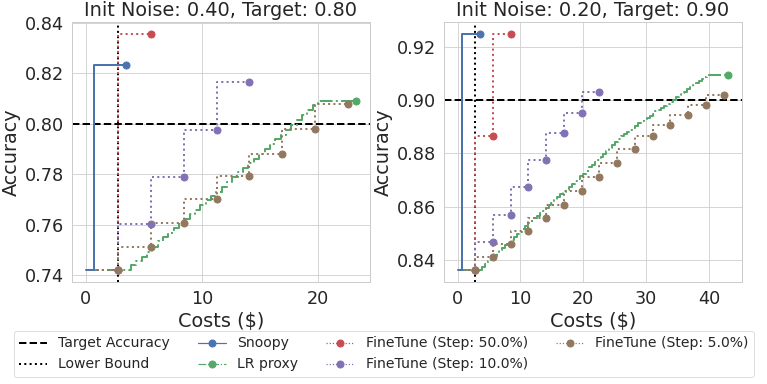}}
\caption{End-to-end use case: (a) CIFAR10 - expensive labels, and (b) IMDB - free labels.}
\label{fig:end2end_cifar10_expensive_imdb_free}
\end{figure*}

\begin{figure*}[ht]
\centering
\subfigure[IMDB - Cheap labels.]{\includegraphics[width=0.45\textwidth]{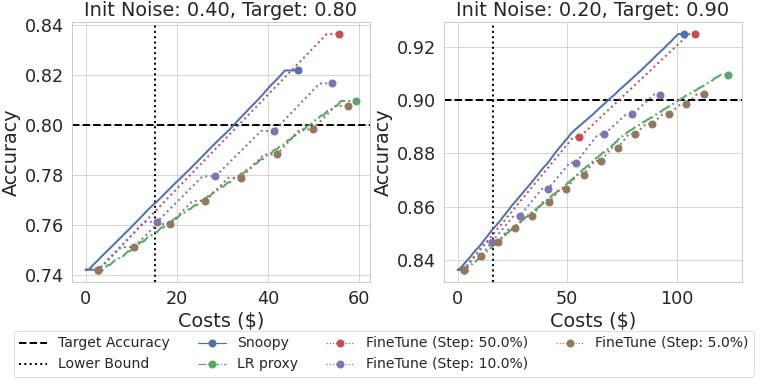}}
\subfigure[IMDB - Expensive labels.]{\includegraphics[width=0.45\textwidth]{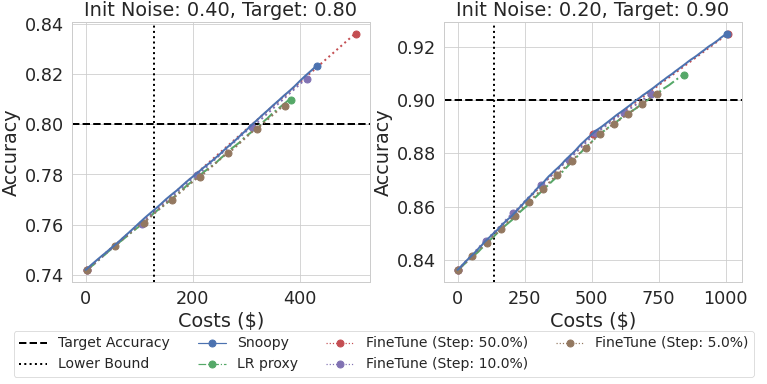}}
\caption{End-to-end use case: (a) IMDB - cheap labels, and (b) IMDB - expensive labels.}
\label{fig:end2end_imdb_cheap_imdb_expensive}
\end{figure*}

\begin{figure*}[ht]
\centering
\subfigure[SST2 - Free labels.]{\includegraphics[width=0.45\textwidth]{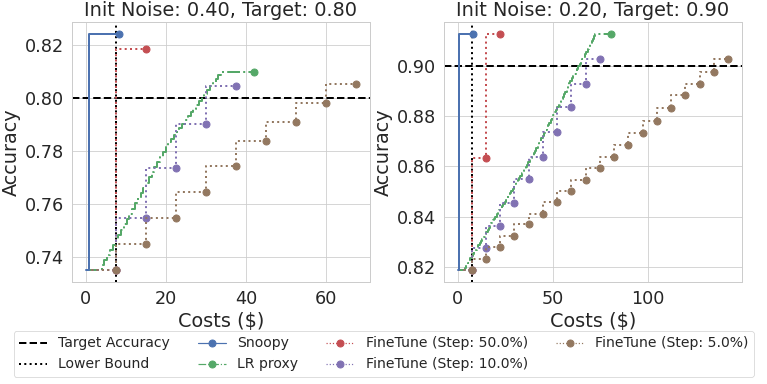}}
\subfigure[SST2 - Cheap labels.]{\includegraphics[width=0.45\textwidth]{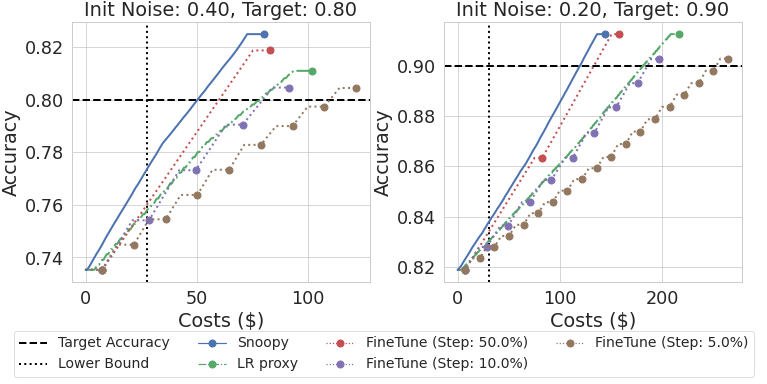}}
\caption{End-to-end use case: (a) SST2 - free labels, and (b) SST2 - cheap labels.}
\label{fig:end2end_sst2_free_sst2_cheap}
\end{figure*}

\begin{figure*}[ht]
\centering
\subfigure[SST2 - Expensive labels.]{\includegraphics[width=0.45\textwidth]{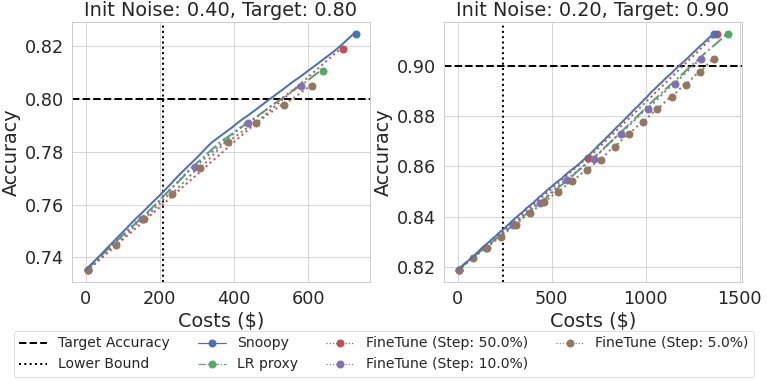}}
\subfigure[YELP - Free labels.]{\includegraphics[width=0.45\textwidth]{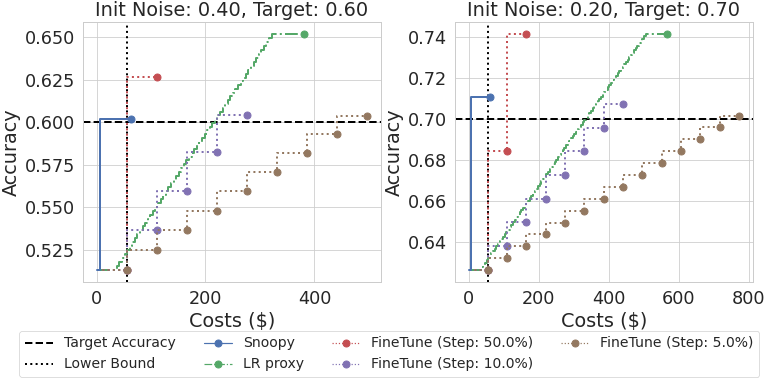}}
\caption{End-to-end use case: (a) SST2 - expensive labels, and (b) YELP - free labels.}
\label{fig:end2end_sst2_expensive_yelp_free}
\end{figure*}

\begin{figure*}[ht]
\centering
\subfigure[YELP - Cheap labels.]{\includegraphics[width=0.45\textwidth]{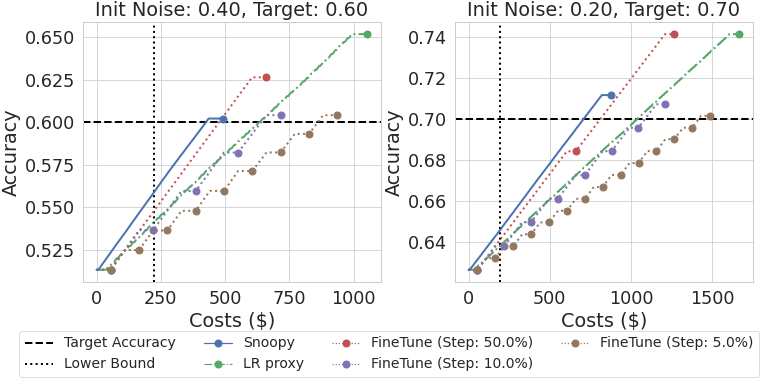}}
\subfigure[YELP - Expensive labels.]{\includegraphics[width=0.45\textwidth]{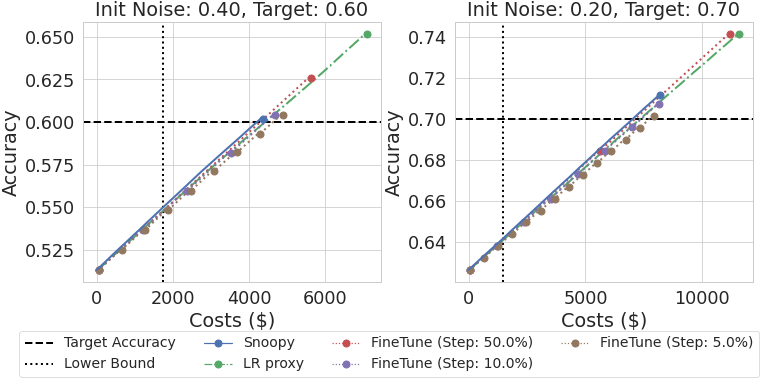}}
\caption{End-to-end use case: (a) YELP - cheap labels, and (b) YELP - expensive labels.}
\label{fig:end2end_yelp_cheap_yelp_expensive}
\end{figure*}

\end{document}